\documentclass[a4paper,fleqn]{cas-sc}

\usepackage[numbers]{natbib}
\usepackage{float}
\usepackage{graphicx}
\usepackage{epstopdf}
\usepackage{subfigure}
\usepackage{lineno}
\usepackage{xcolor}
\usepackage[titletoc,title]{appendix}

\begin{document}
\let\WriteBookmarks\relax
\def\floatpagepagefraction{1}
\def\textpagefraction{.001}
\shorttitle{Physics-integrated neural differentiable modeling for IB flows}
\shortauthors{CL Li, H Xu}

\title [mode = title]{Physics-integrated neural differentiable modeling for immersed boundary systems}

\author[a]{Chenglin Li}[type=editor, auid=000,bioid=1, orcid=0009-0004-3842-4391]
\credit{Conceptualization of this study, Methodology, Software}

\author[a]{Hang Xu}[style=chinese, auid=000, bioid=3, orcid=0000-0003-4176-0738]
\credit{Conceptualization of this study, Draft revision}
\cormark[1]
\cortext[1]{Corresponding author. E-mail address: hangxu@sjtu.edu.cn (H. Xu).}

\address[a]{State Key Lab of Ocean Engineering, School of Ocean and Civil Engineering, Shanghai Jiao Tong University, Shanghai 200240, China }

\author[b]{Jianting Chen}[style=chinese, auid=002, bioid=4]
\credit{b]Draft revision}
\author[b]{Yanfei Zhang}[style=chinese, auid=002, bioid=4]
\credit{Draft revision}

\address[b]{State Key Laboratory of Maritime Technology and Safety, Shanghai Ship and Shipping Research Institute Co., Ltd, Shanghai 200135, China.}


\begin{abstract}
Accurately, efficiently, and stably computing complex fluid flows and their evolution near solid boundaries over long horizons remains challenging. Conventional numerical solvers require fine grids and small time steps to resolve near-wall dynamics, resulting in high computational costs, while purely data-driven surrogate models accumulate rollout errors and lack robustness under extrapolative conditions. To address these issues, this study extends existing neural PDE solvers by developing a physics-integrated differentiable framework for long-horizon prediction of immersed-boundary flows. A key design aspect of the framework includes an important improvement, namely the structural integration of physical principles into an end-to-end differentiable architecture incorporating a PDE-based intermediate velocity module and a multi-direct forcing immersed boundary module, both adhering to the pressure-projection procedure for incompressible flow computation. The computationally expensive pressure projection step is substituted with a learned implicit correction using ConvResNet blocks to reduce cost, and a sub-iteration strategy is introduced to separate the embedded physics module’s stability requirement from the surrogate model’s time step, enabling stable coarse-grid autoregressive rollouts with large effective time increments. The framework uses only single-step supervision for training, eliminating long-horizon backpropagation and reducing training time to under one hour on a single GPU. Evaluations on benchmark cases of flow past a stationary cylinder and a rotationally oscillating cylinder at $Re=100$ show the proposed model consistently outperforms purely data-driven, physics-loss-constrained, and coarse-grid numerical baselines in flow-field fidelity and long-horizon stability, while achieving an approximately 200-fold inference speedup over the high-resolution solver.

\end{abstract}


\maketitle

\section{Introduction}
\noindent 
Complex fluid flows and their evolution in the presence of solid boundaries represent a fundamental class of problems in engineering and natural systems. Despite significant advances in both accuracy and efficiency, conventional computational fluid dynamics (CFD) still suffers from high computational costs, especially when fine resolutions of geometric boundaries and structural dynamic responses are required. This cost becomes particularly prohibitive in flow control and optimization \cite{Rabault_2019}, where large numbers of long-horizon simulations are typically required to identify optimal parameters \cite{COSTANZO2022111664} and control strategies \cite{JIA2025111125, 10.1063/5.0204237}. Hence, developing efficient computational frameworks is of significant importance and has broad practical relevance. 
For flow computations involving complex and/or moving boundaries, the immersed boundary (IB) method \cite{Peskin_2002} solves Eulerian field variables on a fixed Cartesian grid and enforces boundary conditions by correcting near-wall flows using boundary/object information described in a Lagrangian manner. In this way, it avoids the repeated mesh regeneration associated with body-fitted \cite{THOMPSON19821} grids under boundary motion\cite{JOHNSON199473}. The key advantages of IB methods include highly regular computational procedures and data structures \cite{annurev.fluid.37.061903.175743, YANG20125029}, favorable parallel scalability \cite{ZHU2026106913, GIANNENAS2021606}. These properties enable high-throughput time advancement and facilitate embedding boundary-handling modules as operators within an end-to-end computational framework. 

With the development of scientific machine learning, neural networks have been widely used to construct fast surrogate models for fluid \cite{Vinuesa2022} and fluid–structure systems \cite{Davydzenka_Tahmasebi_2022}. Representative approaches include learning spatiotemporal evolution in a latent space via proper orthogonal decomposition (POD) \cite{doi:10.2514/1.J056060} or encoder–decoder architectures\cite{Murata_Fukami_Fukagata_2020}, and learning field mappings on structured/unstructured meshes using convolutional neural networks (CNNs) \cite{Cremades2024, Li_Buzzicotti_2023} or graph neural networks (GNNs) \cite{HADIZADEH2025117921, BARWEY2025117509}. These methods demonstrate the potential of data-driven prediction for high-dimensional spatiotemporal dynamics. However, many purely data-driven black-box models rely heavily on large datasets, and they often suffer from limited generalization under extrapolation in parameter space as well as error accumulation in long-horizon rollouts \cite{annurev-fluid-010719-060214}. Moreover, the cost of data acquisition itself may become a major bottleneck for practical engineering deployment \cite{Liang2022}. 

This has motivated a renewed focus on leveraging accumulated physical priors (governing equations and constraints) to reduce the training difficulty of purely data-driven models \cite{Karniadakis2021}, while exploiting the nonlinear representation capacity of neural networks to alleviate the computational burden of physics-based solvers \cite{10.5555/3305890.3306035}. Achieving this goal requires a careful balance between traditional physics-based frameworks and data-driven surrogates. Existing physics-integrated architectures can be broadly categorized into two lines: (i) incorporating governing equations into the loss functions of deep neural networks (i.e., physics-informed neural networks, PINNs) \cite{Karniadakis2021}, and (ii) embedding physical structure into neural architectures \cite{doi:10.1073/pnas.2101784118} (i.e., PDE-preserved neural networks, referred to as PPNNs \cite{Liu2024}). 

PINNs incorporate physical priors directly into the training objective, typically by regularizing the network with residuals of the governing equations to improve physical consistency and mitigate the lack of constraints in purely data-driven models. Since their introduction, PINNs have been successfully applied across fluid dynamics \cite{JIN2021109951, Zhu_Jiang_2024}, solid mechanics \cite{HAGHIGHAT2021113741}, heat transfer \cite{10.1115/1.4050542}, and related areas. In fluid dynamics, PINNs have been used for solving the Navier–Stokes equations, providing an alternative route for rapid simulation of complex flows. Nevertheless, PINNs usually construct partial differential equation (PDE) residuals via automatic differentiation (AD) \cite{JMLR:v18:17-468} and/or numerical-discretization-based reconstruction, and include them as soft constraints in the loss. This often increases optimization difficulty and makes training highly sensitive to hyperparameters such as the relative weights of loss terms. In high-dimensional parameter spaces or under complex boundary conditions, balancing equation losses and data losses can further lead to training instability or limited gains \cite{NEURIPS2021_df438e52}. 

The second research line, often referred to as PPNNs, embeds physical priors directly into the network architecture through discrete operators. By structurally constraining the hypothesis space, this design reduces the burden on purely data-driven learning and mitigates the accumulation and propagation of nonphysical errors during long-horizon rollouts. The central idea is to establish a mathematical connection between neural-network structure and numerical PDE solvers \cite{Liu2024, NEURIPS2018_69386f6b, pmlr-v80-long18a}. This connection, together with differentiable programming \cite{fan2025diffflowf,BEZGIN2023108527}, enables physical priors to be incorporated into the architecture itself. Such designs provide deeper insights into integrating physics priors with data-driven surrogates. Related works have demonstrated long-time stable computations for complex fluid problems \cite{FAN2024112584, FAN2025117478, AKHARE2023115902}, and have embedded structural-dynamics solvers to enable accurate prediction of structural responses and object boundaries \cite{FAN2024112584}. Although hybrid differentiable neural modeling has shown strong potential, the field remains at an early stage and requires further development, particularly in reducing computational and inference costs for long-horizon prediction \cite{Liu2024, FAN2024112584, Wang_Chu_2025}. 

Based on the aforementioned literature survey, long-horizon prediction for complex immersed boundary flow problems still confronts two core challenges that hinder its practical engineering application, and these challenges motivate the innovations proposed in this work. First, stable long-time forecasts in existing machine learning-based approaches typically rely on sequence-modeling frameworks and multi-step training strategies, which inevitably lead to a significant surge in training cost and time consumption \cite{Wang_Chu_2025, xu2021accelerated}. Moreover, physics-hybrid models, despite their advantages in error mitigation, often suffer from instability at the early training stage and even immediate divergence, posing a critical barrier to their deployment \cite{FAN2024112584}. Second, embedding physical structure into network architectures, while constrains the hypothesis space and reduces effective model capacity, introduces inherent numerical restrictions inherited from traditional physics solvers. If these restrictions are not properly addressed, physical constraints may turn counterproductive by undermining long-horizon stability, slowing down inference speed, and reducing deployment flexibility. This directly conflicts with the core goal of developing efficient computational frameworks for complex flow simulations.

To address the aforementioned challenges, we extend the neural differentiable model introduced by Fan and Wang \cite{FAN2024112584} for long-horizon prediction of flows subjected to immersed boundary systems. Our approach is built upon the PPNNs paradigm \cite{Liu2024}, which establishes an explicit correspondence between the network architecture and numerical PDE discretizations. The work presents four key innovations, each specifically targeting the limitations identified in existing research. For a start, the model architecture and data flow strictly follow the pressure-projection procedure for incompressible flows, ensuring each intermediate variable has a clear physical meaning and thereby significantly mitigating the issue of nonphysical error accumulation \cite{annurev-fluid-010719-060214}. In addition, a sub-iteration strategy decouples the embedded physics solver’s stability constraint from the model time step, which in turn enables stable coarse-grid rollouts with larger time steps to reduce computational overhead \cite{FAN2024112584}. Another key innovation lies in the replacement of the traditional pressure-Poisson projection with a learned correction module, which retains incompressibility constraints while further reducing computational burden \cite{Karniadakis2021, Liang2022}. Lastly, the model achieves long-horizon stable prediction via single-step supervised training, which avoids long-horizon backpropagation to effectively reduce both computational overhead and training complexity \cite{Wang_Chu_2025, xu2021accelerated}.

\section{Case setting} \label{2}

\begin{figure}[ht]
\centering
\includegraphics[scale=.6]{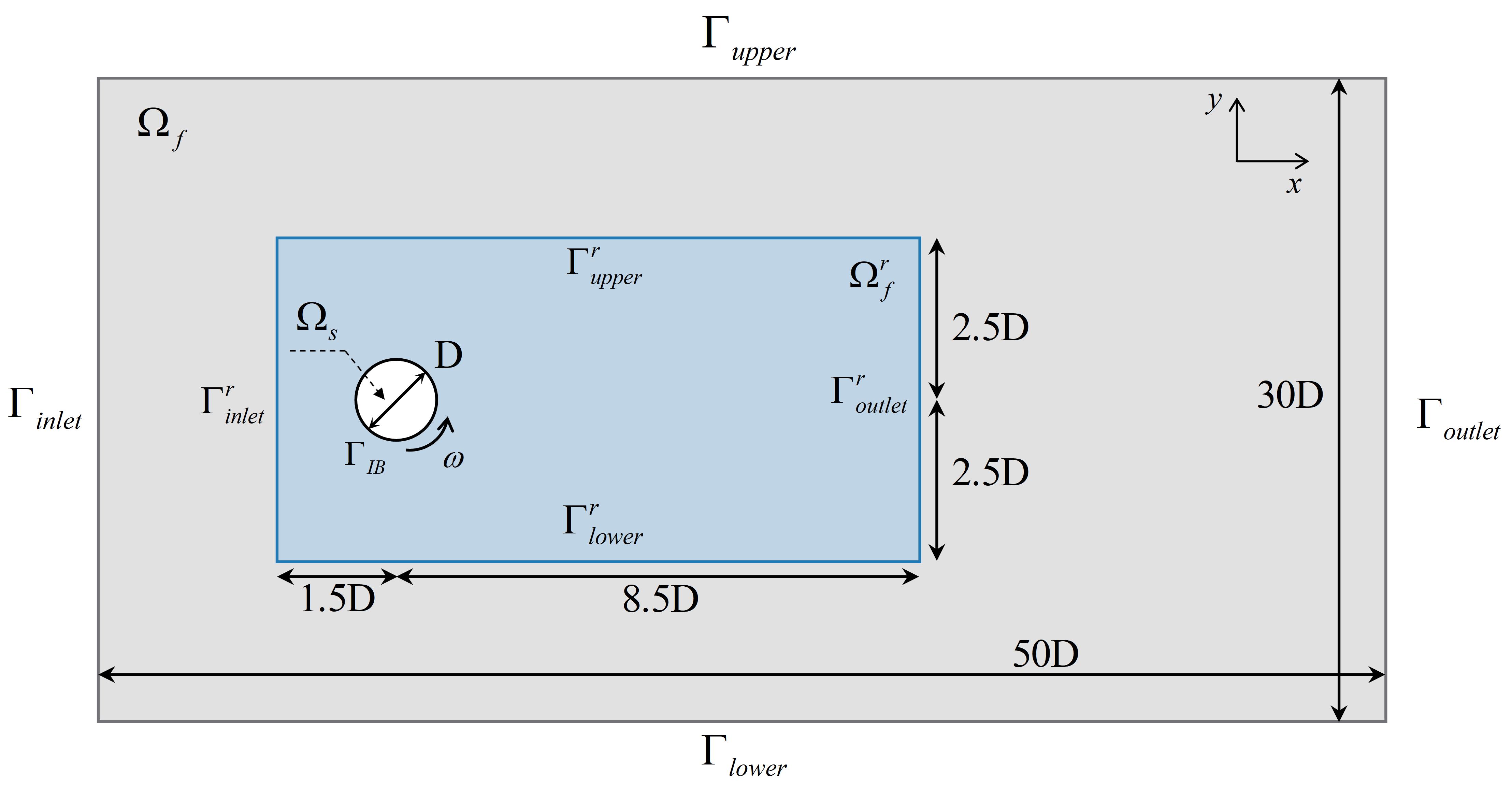}
\caption{Schematic of the computational setup and domains. The immersed-boundary (IB) solver is defined on the global fluid domain $\Omega_f$(gray box) with boundaries $\Gamma_{\mathrm{inlet}}$, $\Gamma_{\mathrm{outlet}}$, $\Gamma_{\mathrm{upper}}$, and $\Gamma_{\mathrm{lower}}$. The surrogate model is evaluated on a truncated subdomain $\Omega^{r}_f$ (blue box) bounded by $\Gamma^{r}_{\mathrm{inlet}}$, $\Gamma^{r}_{\mathrm{outlet}}$, $\Gamma^{r}_{\mathrm{upper}}$, and $\Gamma^{r}_{\mathrm{lower}}$. A prescribed inflow $u=u_{\mathrm{in}}$ is imposed at $\Gamma_{\mathrm{inlet}}$, and free-slip conditions are enforced on $\Gamma_{\mathrm{upper}}$ and $\Gamma_{\mathrm{lower}}$.}
\label{fig001}
\end{figure}

To evaluate the performance of the proposed physics-integrated neural model, we consider two representative IB method benchmark cases: (i) flow past a stationary circular cylinder, and (ii) flow past a circular cylinder subjected to a prescribed rotational oscillation. The computational domain and boundary conditions for both cases are shown in Fig.~\ref{fig001}. Unless otherwise stated, the Reynolds number is fixed at $Re = 100$, with an inflow velocity of $U_\infty = 1\ \text{m/s}$ and a cylinder diameter of $D = 1\ \text{m}$.

First, high-fidelity simulations are conducted on a high-resolution grid using a multi-step direct-forcing IBM scheme; the corresponding numerical verification is provided in the Appendix~\ref{A-}. These simulations serve to construct the high-resolution dataset used as ground truth for both training and testing. In addition, a low-resolution training dataset is obtained by downsampling the high-resolution solutions, with the downsampling procedure detailed in the Appendix~\ref{A.1}.

It is worth noting that, at the classical benchmark condition of $Re=100$, the flow past a stationary cylinder exhibits a pronounced periodic vortex-shedding behavior, which to some extent reduces the difficulty of learning the unsteady flow evolution and the associated hydrodynamic force responses. To provide a more challenging evaluation scenario, we further consider the flow past a cylinder subjected to rotational oscillation (see Fig.~\ref{fig001}), where the prescribed angular velocity is defined as follows:

\begin{equation}
\omega_k=\omega_a \sin\!\left(2\pi f_r k \right),
\end{equation}
where $\omega_a$ denotes the angular-velocity amplitude, $f_r$ is the prescribed (dimensionless) rotation frequency, and $k$ is the $k$-th learning time step. In the present work, we set $f_r=1/5$, i.e., the imposed rotational oscillation has a period of five learning time steps.

Under rotational perturbations, the wake exhibits more intricate spatiotemporal evolution. Meanwhile, the structural load response comprises not only components at the natural frequency but also modulation effects and additional spectral content associated with the imposed rotational-oscillation frequency \cite{Kumar_Lopez2013}, thereby markedly increasing the complexity of the data distribution and the difficulty of learning. Consequently, this benchmark provides a more stringent test of the model’s capability to represent unsteady flows and to generalize across complex dynamical regimes.

\section{Methodology} \label{3}
\subsection{Problem formulation}

The flow around the cylinder is governed by the incompressible Navier–Stokes (NS) equations:
\begin{align}
&\nabla \cdot \mathbf{u} = 0, 
&& (\mathbf{x},t)\in \Omega_f \times [0,T]; \\
&\frac{\partial \mathbf{u}}{\partial t} 
= -(\mathbf{u}\cdot\nabla)\mathbf{u} + \nu \nabla^{2}\mathbf{u} - \frac{1}{\rho}\nabla p,
&& (\mathbf{x},t)\in \Omega_f \times [0,T].
\end{align}
Here $t$ and $\mathbf{x}$ denote time and Eulerian spatial coordinates, respectively. The velocity field $\mathbf{u}(t,\mathbf{x})$ and pressure $p(t,\mathbf{x})$ are the primary spatiotemporal variables in $\Omega_f\subset\mathbb{R}^2$, while $\rho$ and $\nu$ are the fluid density and kinematic viscosity. Given appropriate initial and boundary conditions (IC/BCs), the velocity–pressure solution is well posed and therefore uniquely determined.

To resolve the fluid–solid boundary interaction in this flow system, a multi-direct forcing immersed boundary method (IBM) is employed, where the coupling between the Eulerian (fluid) and Lagrangian (solid) variables is computed via a Dirac delta function. The detailed coupling relations are given as follows:
\begin{align}
&\phi(\mathbf{x},t)=\int_{r_{\mathbf{x}}}\Phi(\mathbf{X},t)\delta(\mathbf{x}-\mathbf{X})d\gamma_{\mathbf{X}},&&\gamma_{\mathbf{X}}\in\Gamma_{IB};\\
&\Phi(\mathbf{X},t)=\int_{r_{\mathbf{x}} }\phi(\mathbf{x},t)\delta(\mathbf{x}-\mathbf{X})d\gamma_{\mathbf{x}},&&\gamma_{\mathbf{x}}\in\Gamma_{IB}.
\end{align}
Here $\Phi$ and $\phi$ are defined on the Lagrangian and Eulerian frameworks, respectively; $\mathbf{x}$ and $\mathbf{X}$ correspond to the set of Eulerian grid nodes and Lagrangian markers in the immersed interface boundary $\Gamma$, respectively, and $\delta$ denotes the Dirac delta interpolation function. In our analysis, a three-point discrete formulation is adopted for this delta function, as given by:
\begin{equation}
\phi(r)=
\begin{cases}
\dfrac{1}{3}\left(1+\sqrt{1-3r^2}\right), & 0 \le r < 0.5, \\[6pt]
\dfrac{1}{6}\left(5-3r-\sqrt{-2+6r-3r^2}\right), & 0.5 \le r < 1.5, \\[6pt]
0, & r \ge 1.5,
\end{cases}
\label{eq:phi}
\end{equation}
where $r = |d|/h$ is the dimensionless normalized distance, $d$ denotes the physical distance between a Lagrangian marker point and an Eulerian grid node along one coordinate direction, and $h$ is the grid spacing.

\subsection{Physics-integrated differentiable neural model}

The proposed IBM-based physics-integrated neural model, as shown in Fig.~\ref{fig002}, is constructed by following the dataflow of a classical pressure-projection solver, with the pressure-correction step performed implicitly by the network module—a key distinction from traditional solvers. The prediction of the intermediate velocity is built upon a principled connection between the neural architecture and the numerical structure of the PDE: as discussed in \cite{Liu2024}, the spatial discretization of differential operators can be interpreted as convolution with fixed kernels. To ensure the numerical stability of the embedded solver, we further adopt a splitting strategy, in which the intermediate-velocity update is carried out via sub-iterations. Details of the fluid and structural solvers are provided as follows:

\begin{figure}[htbp]
\centering
\includegraphics[scale=.7]{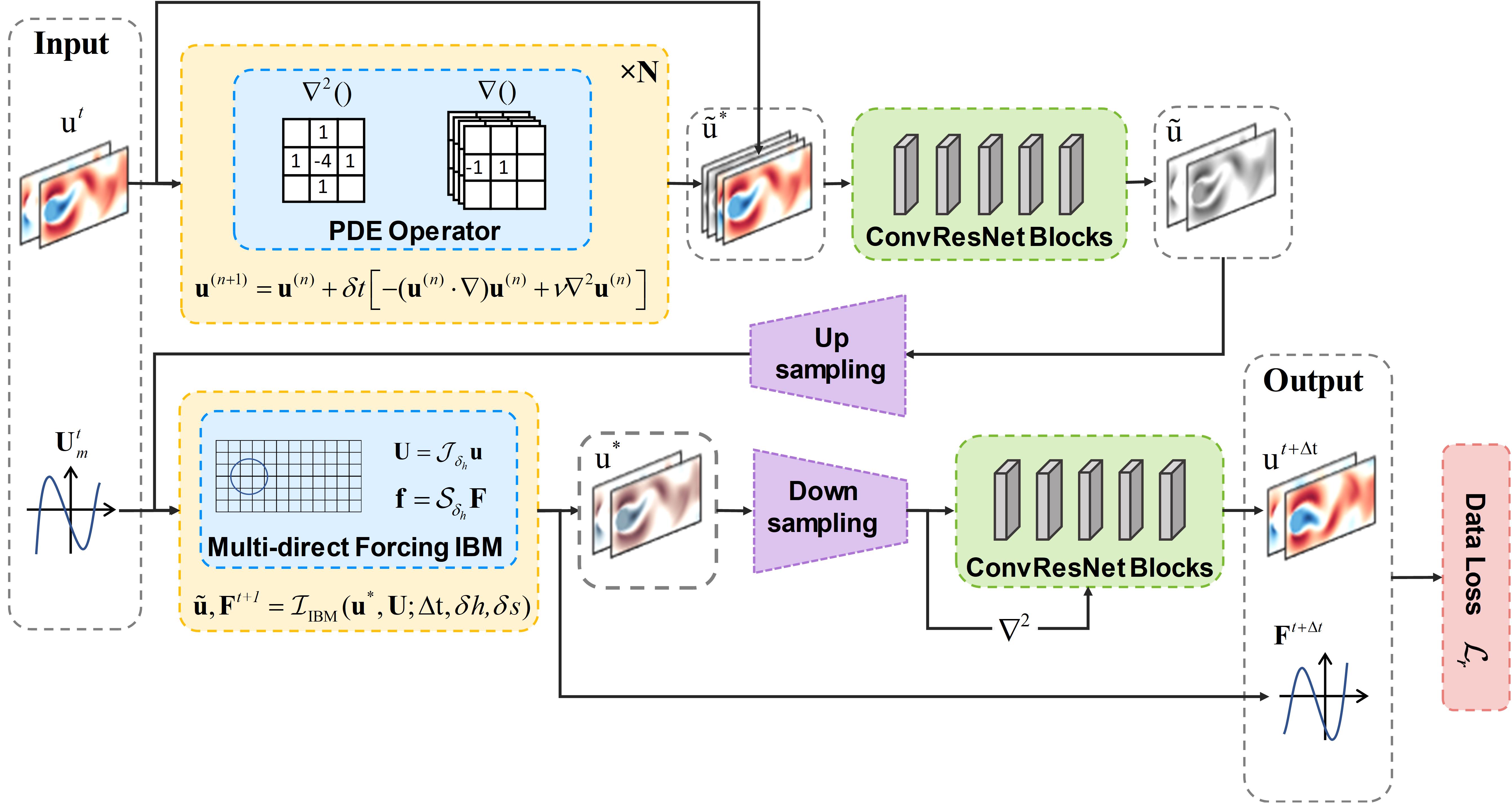}
\caption{Schematic of the proposed physics-integrated differentiable neural network. Trainable neural modules (ConvResNet blocks) are shown in green, while non-trainable physics operators (the PDE operator and the multi-direct-forcing IBM module) are highlighted in yellow. Grey dashed boxes enclose the inputs, outputs, and intermediate physical variables. Data propagate along the black arrows. The PDE-operator substep is iterated $N$ times, and up-/down-sampling bridges the resolutions between the physics and neural components.}
\label{fig002}
\end{figure}

The overall architecture of the model is primarily organized to mirror the workflow of a classical pressure-projection method. Specifically, spatial discretization is encoded via finite differences on a staggered grid, and temporal advancement is performed using a first-order forward Euler scheme. To initiate the velocity prediction process, the intermediate velocity $\tilde{\mathbf{u}}^{*}$ is first computed using the following discrete formulation:
\begin{equation}
\tilde{\mathbf{u}}^{*}=\mathbf{u}^{t}+\Delta t\left[-(\mathbf{u}^{t}\cdot\nabla)\mathbf{u}^{t}+\nu\nabla^{2}\mathbf{u}^{t}\right],
\label{eq:intermediate velocity}
\end{equation}
where $\Delta t$ is the learning time step, which represents the time interval between two consecutive model predictions and is determined by the temporal sampling interval of the training dataset.

Notably, the model’s $\Delta t$ is much larger than the underlying simulation time step $dt$, which far exceeds the CFL stability limit, leading to unreliable solution of the physics-based constraint. This issue results in substantial numerical errors and spurious noise, which severely degrade the accuracy and stability of both training and inference. To address this challenge and enable large-step inference while maintaining numerical robustness, the intermediate-velocity update is reformulated using a splitting strategy. In practice, this strategy is implemented through sub-iterations, with the iteration count $N$ determined according to the CFL constraint. For this work, $N=20$, and accordingly, the update in Eq.~\eqref{eq:intermediate velocity} is rewritten as:
\begin{equation}
\mathbf{u}^{\,n+1}=\mathbf{u}^{\,n}+\delta t\left[-(\mathbf{u}^{\,n}\cdot\nabla)\mathbf{u}^{\,n}+\nu\nabla^{2}\mathbf{u}^{\,n}\right],
\qquad n=0,1,\ldots,N-1,
\end{equation}
where $\delta t=\Delta t/N$ is the sub-iteration time step. In this work, $\Delta t = 0.5$ is determined by the temporal sampling interval of the training dataset, and $N = 20$ sub-iterations are employed such that $\delta t = \Delta t/N = 0.025$ satisfies the CFL constraint on the coarse grid. When $n = 0$, the sub-iteration is initialized with the velocity at the current time level, i.e., $\mathbf{u}^0 = \mathbf{u}^t$; after completing $N$ sub-steps, the intermediate velocity is obtained as $\tilde{\mathbf{u}}^{*} = \mathbf{u}^N$. For spatial discretization, the convective terms are approximated using a first-order upwind scheme, while the diffusive terms are discretized with a second-order central difference scheme. The differential operators are applied using fixed-parameter convolution kernels, ensuring stable and efficient updates for the velocity field during sub-iterations.

This design not only preserves the mathematical stability and physical fidelity of the imposed constraints but also prevents the overall model from being bottlenecked by the numerical stability limit of the physics solver. Under this scheme, $\tilde{\mathbf{u}}^{*}$ is computed by the physics module on a sparse grid via time-step sub-iterations, but it may still contain non-negligible errors and numerical noise. To eliminate these residuals, a trainable ConvResNet module is applied to correct $\tilde{\mathbf{u}}^{*}$, yielding the final intermediate velocity $\tilde{\mathbf{u}}$:
\begin{equation}
\tilde{\mathbf{u}}=\mathcal{F}_{conv_{1}}[\tilde{\mathbf{u}}^{*},\mathbf{u}^{t};\theta_{1}],
\end{equation}
where $\theta_{1}$ is the trainable parameters of the ConvResNet module $\mathcal{F}_{conv_{1}}$.

At this stage, the computed final intermediate velocity $\tilde{\mathbf{u}}$ has not yet incorporated the influence of the immersed solid boundaries within the flow domain. We therefore employ a multi-step direct-forcing immersed boundary (IB) procedure to enforce the no-slip constraint on the immersed interface. Notably, all IB operations are performed on the super-resolved high-resolution fields (the purple modules in Fig.~\ref{fig002}). At the $m$-th IB sub-iteration, the Eulerian intermediate velocity $\tilde{\mathbf{u}}^m$ is first interpolated to the Lagrangian marker locations $\mathbf{X}_K \in \gamma_{\mathbf{X}}$ through the regularized discrete delta kernel, yielding the marker velocity:
\begin{equation}
\tilde{\mathbf{U}}^{\,m}(\mathbf{X}_K)=\sum_{\mathbf{x}_i\in\gamma_\mathbf{x}}\tilde{\mathbf{u}}^{\,m}(\mathbf{x}_i)\delta_h(\mathbf{x}_i-\mathbf{X}_K)h^2,\quad\mathbf{X}_K\in\gamma_\mathbf{X},\qquad  m=0,1,\ldots,M-1,
\label{eq:ibm begin}
\end{equation}
where $\tilde{\mathbf{U}}^m(\mathbf{X}_K)$ denotes the interpolated velocity at the $K$-th marker, $h$ is the uniform grid spacing, and $\delta_h(\cdot)$ is the regularized discrete delta kernel. This kernel is constructed as the two-dimensional tensor product of the one-dimensional kernel $\phi(r)$ defined in Eq.~\eqref{eq:phi}, specifically:
\begin{equation}
\delta_h(\mathbf{x}_i-\mathbf{X}_K)
=\frac{1}{h^2}\,
\phi\!\left(\frac{\bigl|x_i^{(1)}-X_K^{(1)}\bigr|}{h}\right)
\phi\!\left(\frac{\bigl|x_i^{(2)}-X_K^{(2)}\bigr|}{h}\right),
\label{eq:delta_h}
\end{equation}
where $x_i^{(1)}$ and $x_i^{(2)}$ denote the two Cartesian components of the Eulerian node $\mathbf{x}_i$, and $X_K^{(1)}$ and $X_K^{(2)}$ represent the two Cartesian components of the Lagrangian marker $\mathbf{X}_K$. The factor $1/h^2$ ensures that the discrete kernel approximates the two-dimensional Dirac delta distribution in the continuum limit. Based on the mismatch between the interpolated velocity $\tilde{\mathbf{U}}^m_K$ and the prescribed target boundary velocity $\mathbf{U}_K^{target}$, the direct-forcing term on each marker is defined as:
\begin{equation}
\mathbf{F}^{\,m}_K
=
\frac{\mathbf{U}^{\,target}_K-\tilde{\mathbf{U}}^{\,m}_K}{\Delta t},
\qquad \mathbf{X}_K\in\gamma_{\mathbf{X}},\qquad  m=0,1,\ldots,M-1,
\end{equation}
where $\Delta t$ is the learning time step (i.e., the model's prediction interval), which is consistent with the global step defined in Eq.~\eqref{eq:intermediate velocity}. This consistency is enabled by our sub-iteration design, which decouples the overall model from the restrictive stability constraint associated with a small simulation time step. After determining the marker forcing, it is subsequently spread back to the Eulerian grid to form a body-force density:
\begin{equation}
\mathbf{f}^m(\mathbf{x}_i)=\sum_{\mathbf{X}_K\in\gamma_\mathbf{X}}\mathbf{F}^{\,m}_K\delta_h(\mathbf{x}_i-\mathbf{X}_K)\Delta s_K,
\qquad \mathbf{x}_i\in\gamma_{\mathbf{x}},
\end{equation}
with $\Delta s_K = h\,\delta s$ denoting the discrete boundary measure associated with marker $K$, where $\delta s$ is the mean spacing between adjacent Lagrangian markers. Using this body-force density, the Eulerian velocity field is then corrected accordingly:
\begin{equation} 
\tilde{\mathbf{u}}^{m+1}(\mathbf{x}_i)=\tilde{\mathbf{u}}^m(\mathbf{x}_i)+\Delta t\mathbf{f}^m(\mathbf{x}_i),\qquad m=0,1,\ldots,M-1,
\label{eq:ibm end}
\end{equation}
The above gather-force-spread-update cycle (from Eq.~\ref{eq:ibm begin} to Eq.~\ref{eq:ibm end}) is repeated for $M$ sub-iterations to ensure the no-slip constraint is fully enforced. In this work, $M = 5$ sub-iterations are employed, which is found to be sufficient for the no-slip constraint to converge on the immersed boundary. Upon completion of the IB sub-iterations, the boundary-corrected final intermediate velocity is obtained as:
\begin{equation}
\mathbf{u}^{*}=\tilde{\mathbf{u}}^{M},
\end{equation}
which serves as the intermediate velocity after enforcing the immersed-boundary constraints.

Beyond correcting the velocity field, the immersed-boundary procedure also provides a direct route to evaluating the hydrodynamic force exerted on the immersed body. At each sub-iteration $m$, the direct-force term $\mathbf{F}_{K}^{m}$ represents the momentum source per unit time required to satisfy the no-slip condition at marker $K$. Summing the equal-and-opposite reaction over all markers and accumulating the contributions across all $M$ sub-iterations yields the net hydrodynamic force at the advanced time level:
\begin{equation}
\mathbf{F}(t+\Delta t)=\sum_K\left(-\sum_{m=0}^{M-1}\mathbf{F}_K^m\Delta s_K\right),
\end{equation}
where the negative sign follows from Newton's third law, and the outer summation over sub-iterations accounts for the cumulative momentum imparted to the fluid through successive velocity corrections (Eqs.~\ref{eq:ibm begin}--\ref{eq:ibm end}). This force-evaluation strategy is consistent with the multi-direct-forcing formulation of~\cite{WANG2008283}.

Subsequently, a pressure-correction step is applied to $\mathbf{u}^*$ to recover a divergence-free velocity consistent with the incompressibility constraint of the NS equations. In a conventional projection framework, this correction is obtained by solving a pressure Poisson equation—an elliptic problem that requires repeated solutions of a large sparse linear system and typically dominates the computational cost of incompressible CFD. In the proposed architecture (Fig.~\ref{fig002}), this bottleneck is avoided by introducing a ResNet-based implicit pressure-correction module. Specifically, the divergence field $\nabla \cdot \mathbf{u}^*$ (together with the intermediate velocity features) is provided as input to the ConvResNet blocks, which learn to predict the pressure-induced correction and directly output the corrected velocity at the next time level:
\begin{equation}
\mathbf{u}(t+\Delta t)=\mathcal{F}_{conv_{2}}\left[\mathbf{u}^{*}(t),\,\nabla \cdot \mathbf{u}^*(t);\,\theta_{2}\right],
\end{equation}
where $\theta_{2}$ is the trainable parameters of the ConvResNet module $\mathcal{F}_{conv_{2}}$. As a result, the model enforces the projection effect without explicitly solving the pressure Poisson system, yielding an efficient end-to-end differentiable surrogate while retaining the essential incompressibility constraint.

\subsection{Single-step model training}

Previous studies have achieved long-horizon stable inference using backpropagation through time (BPTT)-based training schemes\cite{58337}; however, such strategies are not well suited to physics-integrated models. As reported in \cite{FAN2024112584}, BPTT training can be unstable in the early stages; moreover, because the physics module is substantially more computationally demanding than the neural network component, BPTT training typically incurs considerable GPU memory overhead. In contrast, the proposed differentiable physics-integrated network is trained using a one-step supervised strategy, where the model is optimized to match the ground-truth flow fields and hydrodynamic forces at the next time level without temporal unrolling.

Given the input state at time $t$, the model predicts the velocity components and the resultant force at $t+\Delta t$, denoted by $\hat{u}^{t+\Delta t},\hat{v}^{t+\Delta t}$, and $\hat{\mathbf{F}}^{t+\Delta t}$, respectively. The overall training objective is defined as:
\begin{equation}
\mathcal{L}(\theta_1, \theta_2)
=
\lambda_{\mathbf{u}}(\mathcal{L}_{u}(\theta_1, \theta_2)
+
\mathcal{L}_{v}(\theta_1, \theta_2))
+
\lambda_{F}\,\mathcal{L}_{\mathbf{F}}(\theta_1),
\label{eq:loss_total_onestep}
\end{equation}
where $\lambda_\mathbf{u}$ and $\lambda_\mathbf{F}$ weight the contributions of the velocity and force supervision terms, respectively. Here we simply set $\lambda_{\mathbf{u}}=\lambda_\mathbf{F}=1$ for all experiments to balance the two supervision terms.

Conventional data-normalization heuristics do not, in general, conform to the principle of physical similarity (i.e., the scaling implied by similarity laws), which may lead to inconsistent error metrics across heterogeneous targets. To ensure a consistent metric across velocity and force predictions, we adopt a relative $L_p$ error (with $p = 2$ in this work) as the basic loss functional. For any prediction–target pair $(\hat{y}, y)$, the relative loss is defined as:
\begin{equation}
\mathcal{R}_{p}(\hat{\mathbf{y}},\mathbf{y})
=
\frac{\|\hat{\mathbf{y}}-\mathbf{y}\|_{p}}{\|\mathbf{y}\|_{p}+\varepsilon},
\label{eq:rel_Lp}
\end{equation}
where $\| \cdot \|_{p}$ denotes the $L_{p}$ norm computed over all degrees of freedom (i.e., after flattening the corresponding field), and $\varepsilon$ is a small constant for numerical stability (set to $10^{-8}$ in this paper) to avoid division by zero. Based on this relative $L_p$ error, the three loss components are compactly written as:
\begin{equation}
\begin{aligned}
\mathcal{L}_{u}(\theta_1, \theta_2)
&=
\frac{1}{|\mathcal{B}|}\sum_{b\in\mathcal{B}}
\mathcal{R}_{p}\!\left(\hat{u}^{\,t+\Delta t}_{b},\,u^{\,t+\Delta t}_{b}\right),\\
\mathcal{L}_{v}(\theta_1, \theta_2)
&=
\frac{1}{|\mathcal{B}|}\sum_{b\in\mathcal{B}}
\mathcal{R}_{p}\!\left(\hat{v}^{\,t+\Delta t}_{b},\,v^{\,t+\Delta t}_{b}\right)\\
\mathcal{L}_{\mathbf{F}}(\theta_1)
&=
\frac{1}{|\mathcal{B}|}\sum_{b\in\mathcal{B}}
\mathcal{R}_{p}\!\left(\hat{\mathbf{F}}^{\,t+\Delta t}_{b},\,\mathbf{F}^{\,t+\Delta t}_{b}\right),
\end{aligned},
\label{eq:loss_components_relLp}
\end{equation}
where $\mathcal{B}$ denotes the mini-batch index set. Here, $\mathbf{F}^{t+\Delta t} = [F_x^{t+\Delta t}, F_y^{t+\Delta t}]^\top$ denotes the ground-truth force vector and $\hat{\mathbf{F}}^{t+\Delta t}$ is the corresponding model prediction.

Finally, the model parameters are obtained by minimizing the total loss over the training set via stochastic gradient descent:
\begin{equation}
(\theta^{*}_1,\theta^{*}_2)=\arg\min_{\theta_1,\theta_2}\left[\mathcal{L}(\theta_1,\theta_2)\right].
\label{eq:loss_opt_onestep}
\end{equation}
Because the optimization is performed on single-step targets, the training does not require backpropagation through long rollouts, which substantially reduces memory consumption and improves training robustness for physics-integrated architectures.

\section{Result and analysis} \label{4}
\subsection{Baseline models for comparison}

To better assess and benchmark the advantages of the proposed model, we consider two widely used classes of learning-based approaches: (i) a purely data-driven model and (ii) a physical loss-constrained model with additional physical loss terms. Schematic diagrams of these architectures are provided in Appendix~\ref{B}. In addition, we include a purely numerical baseline solved on the same coarse grid and compare its results against the learning-based models, ensuring a comprehensive and fair evaluation framework.

The purely data-driven network, as shown in Fig.~\ref{baseline model}(a), is constructed by removing all physics-based computation steps from our proposed model. Specifically, the fluid update is implemented using a ConvResNet module with the same architecture as that employed in our model. For the structural dynamics component, to preserve a consistent data flow, the original IB method computation block is replaced by an equivalent ConvResNet module, while all other components remain unchanged. The total number of trainable parameters is kept comparable to that of our model, and we adopt the same training strategy and loss functions to eliminate confounding variables. Overall, the purely data-driven baseline consists of three independent ResNet networks; its data flow and training protocol are strictly aligned with those of our model, thereby enabling a fair and controlled comparison of performance.

For the physical loss-constrained model, as shown in Fig.~\ref{baseline model}(b), the network architecture is identical to that of the purely data-driven baseline. The only distinction lies in its loss function, which comprises a physics-based loss term and a data-fitting loss term, defined as follows:
\begin{equation}
\mathcal{L}_{\mathrm{total}}
=
\alpha\,\mathcal{L}_{\mathrm{phy}}
+
\beta\,\mathcal{L}_{r},
\label{eq:loss_total_phy}
\end{equation}
where $\alpha$ and $\beta$ are weighting coefficients selected following the approach in \cite{Liu2024} to ensure consistent physical regularization strength.

The data loss $\mathcal{L}_{r}$ is kept identical to that used in the other models to ensure a fair comparison; the physics-consistency term $\mathcal{L}_{\mathrm{phy}}$ is specified as follows:
\begin{equation}
\mathcal{L}_{\mathrm{phy}}
=
\mathcal{L}_{m}
+
\mathcal{L}_{\mathrm{div}}
+
\mathcal{L}_{\mathrm{IB}},
\label{eq:loss_phy_decomp}
\end{equation}
where $\mathcal{L}_m$ and $\mathcal{L}_{\mathrm{div}}$ correspond to the momentum and continuity residual penalties, respectively, and $\mathcal{L}_{\mathrm{IB}}$ accounts for the boundary-condition consistency at the immersed interface.

\begin{equation}
\begin{aligned}
&\mathcal{L}_{m}
=
\left\langle
\left\|
\frac{\partial \mathbf{u}}{\partial t}
+
(\mathbf{u}\cdot\nabla)\mathbf{u}
-
\nu\nabla^2\mathbf{u}
\right\|_2^2
\right\rangle,
\qquad
\mathbf{x}\in \Omega_f;
\\
&\mathcal{L}_{\mathrm{div}}
=
\left\langle
\left|
\nabla\cdot\mathbf{u}
\right|^2
\right\rangle,
\qquad
\mathbf{x}\in \Omega_f;
\\
&\mathcal{L}_{\mathrm{IB}}
=
\left\langle
\mathcal{R}_{p}(
\mathbf{u}_{\Gamma}
,
\mathbf{u}^{\,\mathrm{bc}}
)
\right\rangle,
\qquad
\mathbf{x}\in \Gamma.
\end{aligned}
\end{equation}
Here, $\langle\cdot\rangle$ denotes averaging over the training batch and the corresponding spatiotemporal sample points (or grid locations), and $\Gamma$ denotes the boundary (and/or IB constraint) locations. In practice, $\mathcal{L}_{m}$ and $\mathcal{L}_{\mathrm{div}}$ are evaluated using a scheme-consistent discrete approximation that matches the numerical update employed in the solver, ensuring that the physical loss aligns with the underlying PDE discretization.

\subsection{Flow past a stationary circular cylinder} \label{4.2}
To evaluate the long-horizon rollout capability of the proposed model, we construct the training set over the nondimensional time window $t^*\in[0,100]$, where $t^*=tU_\infty/D$, and adopt the same one-step supervised training strategy as described in Section~\ref{3}. After training, the model is initialized at $t^*=0$ and iteratively rolled out until $t^*=200$. Accordingly, predictions over $t^*\in[0,100]$ correspond to rollouts within the training horizon, whereas those over $t^*\in[100,200]$ assess the model's extrapolation performance beyond the trained time range. The training hyperparameters are summarized in Appendix~\ref{A.2} for reproducibility.

\begin{figure}[htbp]
\centering
\includegraphics[scale=.6]{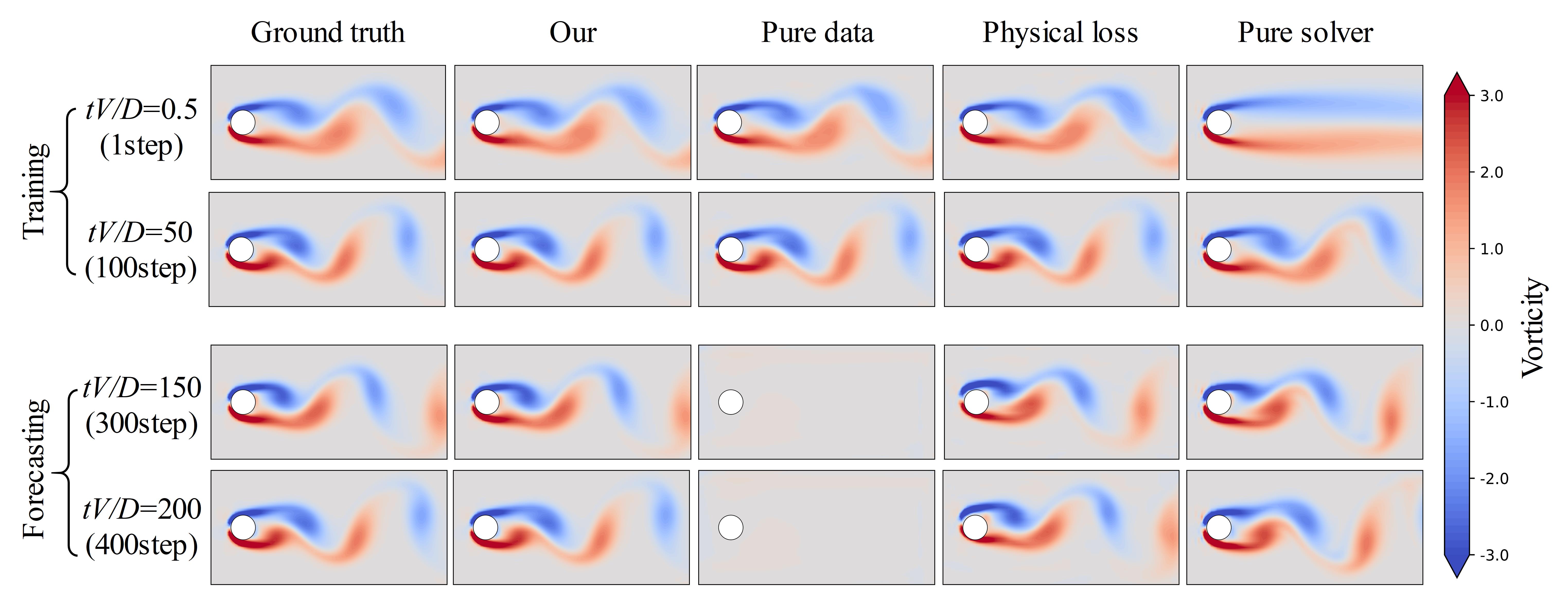}
\caption{Comparison of spatio-temporal vorticity predictions by different neural models for flow past a stationary circular cylinder at $(Re=100, \omega=0)$.}
\label{r0_vortex}
\end{figure}

We compare the proposed method against other baseline models from both spatial and temporal perspectives, as shown in Fig.~\ref{r0_vortex} and Fig.~\ref{force_r0}. Specifically, the vorticity snapshots assess the models’ capability to infer complex spatial flow structures, whereas the time histories of structural loads evaluate their long-horizon rollout stability and accuracy. The results indicate that the proposed model consistently outperforms all baseline models in both aspects, even though it is trained under a one-step supervised strategy using only non-sequential features—highlighting the advantage of embedding physical constraints into the model architecture.

\begin{figure}[htbp]
\centering
\includegraphics[scale=.35]{figure/force_r0.jpg}
\caption{Comparison of structural force predictions by different neural models for flow past a stationary circular cylinder at $(Re=100, \omega=0)$. The left column shows the time histories of the drag coefficient $C_D$, and the right column shows the lift coefficient $C_L$. The gray-shaded interval denotes the training window, whereas the white interval corresponds to out-of-distribution time extrapolation beyond the training horizon. The black solid line represents the ground truth, and the blue curve corresponds to the prediction of the proposed model.}
\label{force_r0}
\end{figure}

By benchmarking against the high-resolution numerical solution (the fifth column in Fig.~\ref{r0_vortex}), we observe that simulations at coarse resolution suffer from pronounced numerical dissipation, leading to substantial prediction errors. Moreover, the excessive numerical viscosity significantly delays vortex shedding relative to the high-resolution reference, demonstrating the limitations of coarse-grid numerical solvers for accurate flow prediction. In contrast, the purely data-driven model diverges rapidly after only a short rollout horizon, which differs from the findings reported in \cite{FAN2024112584}. This discrepancy mainly arises because, in the absence of embedded physical constraints, the model learns a purely data-driven one-step evolution map and is then directly exposed to long-horizon autoregressive rollout under single-step supervision. Under this setting, even small local prediction errors are recursively fed back into subsequent inputs, causing progressive drift of the learned dynamics and eventual breakdown. More importantly, such black-box autoregressive models do not explicitly enforce the governing evolution structure or the immersed-boundary interaction, so the rollout can easily depart from the physically admissible solution manifold once noticeable errors appear. For this reason, many existing CFD surrogate models\cite{Hasegawa_2020, 10.1063/5.0210493} rely on explicitly temporal architectures (e.g., RNNs or Transformers) together with BPTT training schemes to partially suppress error accumulation, albeit at the cost of substantially increased data requirements and training time.

After incorporating the physics-based loss term, the long-horizon prediction accuracy of flow evolution is markedly improved, consistent with prior research\cite{10.1063/5.0312013}. Nevertheless, comparison with the ground truth reveals a persistent phase drift after prolonged rollouts, which is even more evident in structural force prediction (the third row in Fig.~\ref{force_r0}). This behavior indicates that soft physics regularization alone is insufficient to prevent cumulative dynamical deviation: although the loss term encourages physical consistency during training, it does not explicitly constrain the rollout trajectory to follow the numerical update process or strictly preserve the fluid–solid coupling at the solid boundary. As a result, boundary-condition mismatch and phase error can still accumulate over time, eventually degrading both wake prediction and force estimation.

In contrast, the proposed model accurately captures both the near-cylinder flow field and the wake dynamics from the onset of vortex shedding to its fully developed regime. This advantage is not merely empirical, but stems directly from the model design. First, the discrete PDE operators are embedded into the residual update, so the network no longer learns an unconstrained black-box step-to-step mapping; instead, its evolution is restricted by the known numerical physics, which effectively reduces the admissible solution space and suppresses the emergence of spurious dynamics during long rollouts. Second, the multi-resolution mechanism improves the accuracy of the immersed boundary forcing, enabling more faithful resolution of the fluid–solid interaction and better enforcement of the boundary effect near the cylinder surface. This is particularly important in the present problem, where long-term wake development and force response are highly sensitive to the accuracy of near-wall flow evolution. Consequently, the proposed model shows substantially weaker error accumulation than both the purely data-driven model and the physics-loss-constrained model.

A similar trend is observed for structural load prediction, as shown in the first row in Fig.~\ref{force_r0}. Although the early-stage dynamics fall within the training rollout range, the flow field in this regime remains highly transitional and the vortex structures are still forming, making the force response especially sensitive to small phase and boundary errors. As a result, both the purely data-driven model and the physics-loss-constrained model already accumulate large errors at this stage. Benefiting from the explicit physical update path and the more accurate treatment of immersed-boundary interaction, the proposed model maintains both amplitude and phase consistency over long horizons, producing structural load predictions that agree closely with the reference solution.

\begin{figure}[htbp]
\centering
\includegraphics[scale=.7]{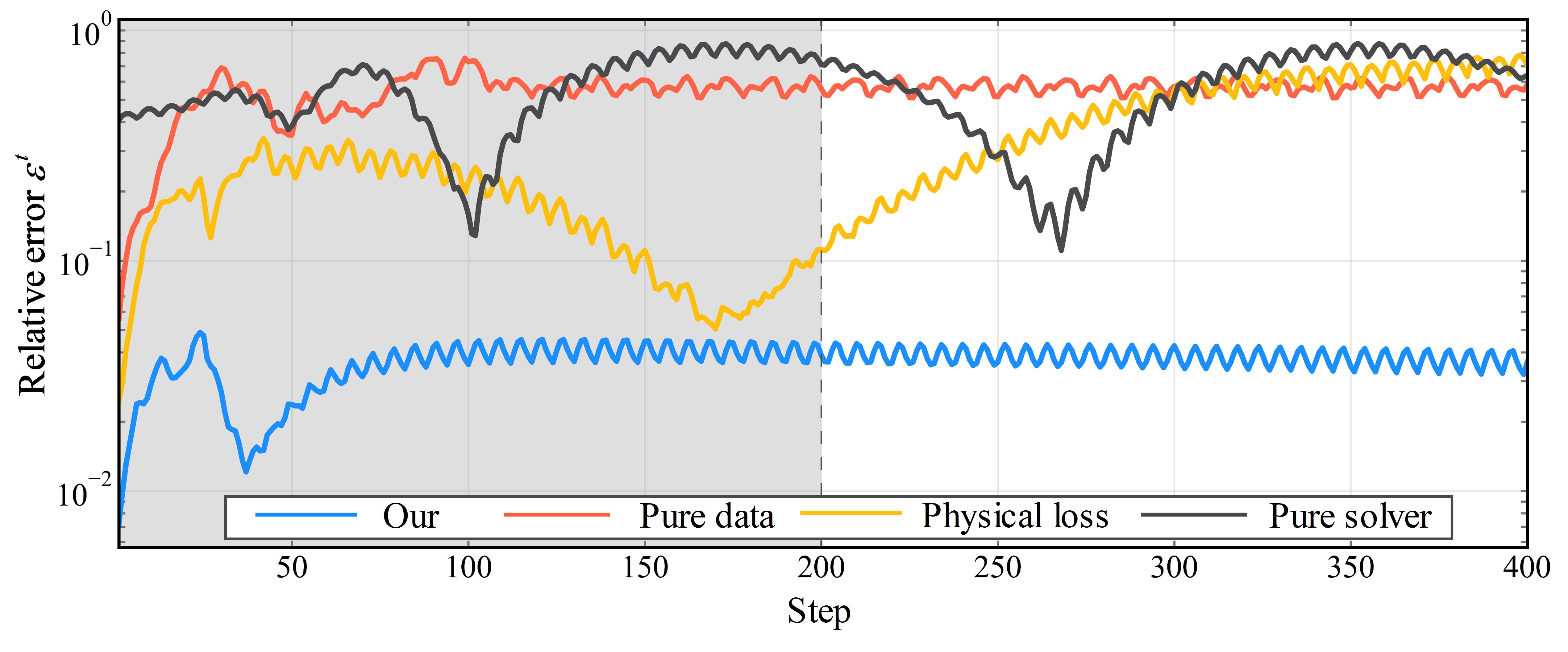}
\caption{Stepwise rollout relative error $\epsilon^{\,t}$ for flow past a stationary circular cylinder at $Re=100$ and $\omega=0$, defined in Eq.~\eqref{eq:step_rel_error_t}. The metric is computed as the average of the relative $L_2$ errors of $u$, $v$, and the structural response vector $\mathbf{F}$ at each rollout step. The gray-shaded region indicates the training window, and the white region indicates temporal extrapolation beyond the training horizon.}
\label{r0_rerror}
\end{figure}

To facilitate a clearer comparison of the long-horizon rollout performance across different models, the relative error at each time step is defined as follows:
\begin{equation}
\epsilon^{\,t}
=
\frac{1}{3}
\Big(
\mathcal{R}_{2}\!\left(\hat{u}^{\,t},u^{\,t}\right)
+
\mathcal{R}_{2}\!\left(\hat{v}^{\,t},v^{\,t}\right)
+
\mathcal{R}_{2}\!\left(\hat{\mathbf{F}}^{\,t},\mathbf{F}^{\,t}\right)
\Big),
\label{eq:step_rel_error_t}
\end{equation}
where $t$ denotes the time index of the rollout, $\hat{u}^{\,t}$ and $\hat{v}^{\,t}$ are the predicted horizontal and vertical velocity components at time $t$, and $u^{\,t}$ and $v^{\,t}$ are the corresponding reference solutions. $\hat{\mathbf{F}}^{\,t}$ and $\mathbf{F}^{\,t}$ denote the predicted and reference structural response vectors, respectively. The operator $\mathcal{R}_{2}(\cdot,\cdot)$ is the relative $L_{2}$ error defined in Eq.~\eqref{eq:rel_Lp}. The velocity error is evaluated separately for the two components and then averaged, whereas the structural force error is computed using the full response vector as a whole. As shown in Fig.~\ref{r0_rerror}, the purely numerical solver (black) exhibits a relative error that is several orders of magnitude higher from the very beginning, primarily due to numerical viscosity and discretization errors. This observation further confirms that, at coarse resolution, relying solely on low-fidelity numerical solvers for accelerated computation is no longer effective. The purely data-driven model (red) accumulates errors rapidly at the early stage of the rollout and remains at a relatively high level thereafter. The physics loss-constrained model (yellow) also shows rapid initial error growth; the oscillations mainly stem from the phase mismatch between the predictions and the reference solution. In contrast, the proposed model (blue) achieves substantially smaller errors than the aforementioned baselines throughout the rollout and effectively suppresses error accumulation over long-horizon inference.

\subsection{Flow past a rotationally oscillating circular cylinder} \label{4.3}

To rigorously assess the performance of the proposed model and to mitigate potential bias arising from a single test scenario, we further evaluate the model in the rotationally oscillating cylinder configuration. The case setup follows that of the benchmark in Section~\ref{2}, ensuring consistency with the overall problem formulation. To better highlight the capability of the proposed approach, we consider an angular-velocity amplitude of $\omega=4~\mathrm{rad/s}$, corresponding to a non-lock-in regime. This regime is particularly challenging because the wake dynamics are no longer governed by a single dominant shedding frequency; instead, they contain both intrinsic vortex-shedding components and rotation-induced disturbances, resulting in weaker periodicity and more intricate fluid–structure interaction. Consequently, this case provides a more stringent test of whether a surrogate model can preserve physically consistent long-horizon evolution beyond simple periodic extrapolation. The training hyperparameters for all models are kept identical to those in Section~\ref{4.2} to ensure a fair comparison.

\begin{figure}[htbp]
\centering
\includegraphics[scale=.6]{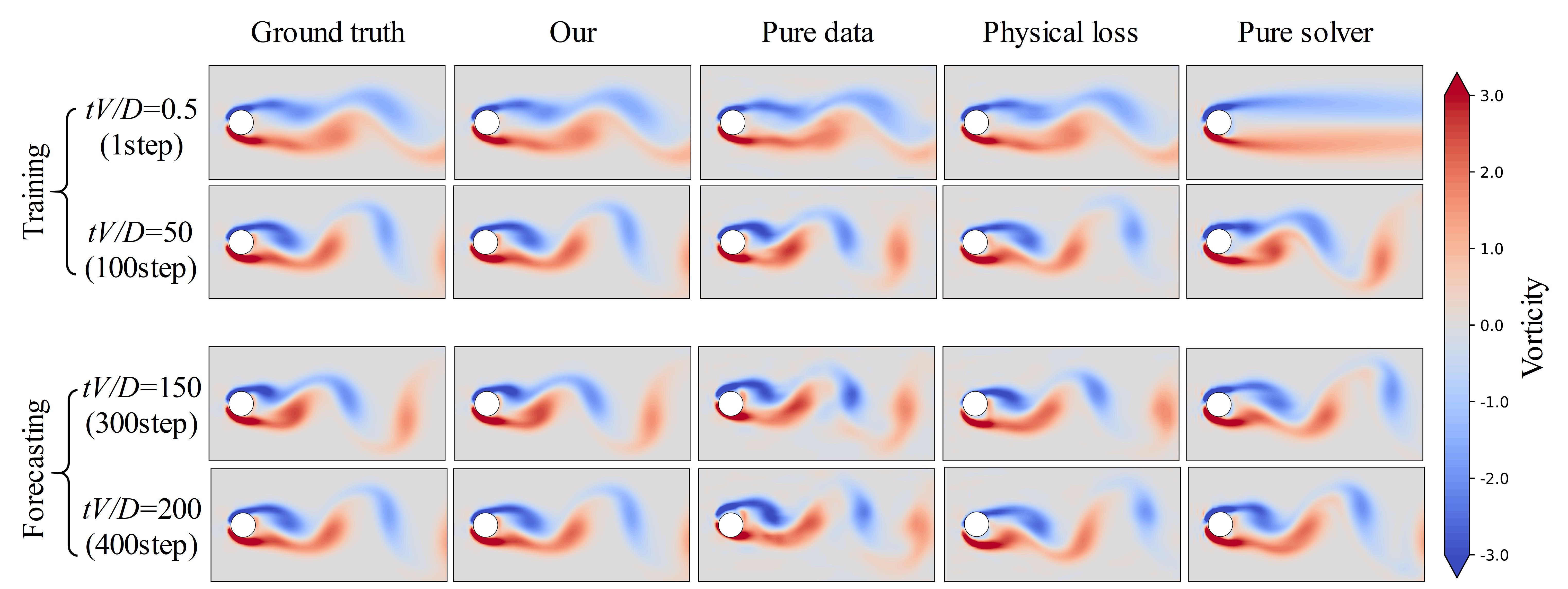}
\caption{Comparison of spatio-temporal vorticity predictions by different neural models for flow past a rotationally oscillating circular cylinder at $(Re=100, \omega=4)$.}
\label{r4_vortex}
\end{figure}

The evaluation results are presented in Figs.~\ref{r4_vortex} and \ref{force_r4}, and the conclusions are largely consistent with those reported in Section~\ref{4.2}. At coarse resolution, the purely numerical solver still substantially overestimates the structural loads and exhibits considerable errors in predicting the vortex-shedding frequency. As the rollout horizon increases, both the purely data-driven model and the physics-loss-constrained model exhibit clear drift of the solid region, accompanied by progressively amplified phase deviations in the wake vortices. These errors are more critical in the present non-lock-in regime, where the flow and force responses are less regular and therefore more sensitive to small inaccuracies in boundary treatment and temporal evolution. Once such errors emerge, they readily distort the coupled wake development and structural loading, leading to cumulative degradation in long-horizon predictions.

\begin{figure}[htbp]
\centering
\includegraphics[scale=.35]{figure/force_r4.jpg}
\caption{Comparison of structural force predictions by different neural models for flow past a rotationally oscillating circular cylinder at $(Re=100, \omega=4)$. The left column shows the time histories of the drag coefficient $C_D$, and the right column shows the lift coefficient $C_L$. The gray-shaded interval denotes the training window, whereas the white interval corresponds to out-of-distribution time extrapolation beyond the training horizon. The black solid line represents the ground truth, and the blue curve corresponds to the prediction of the proposed model.}
\label{force_r4}
\end{figure}

\begin{figure}[htbp]
\centering
\includegraphics[scale=.7]{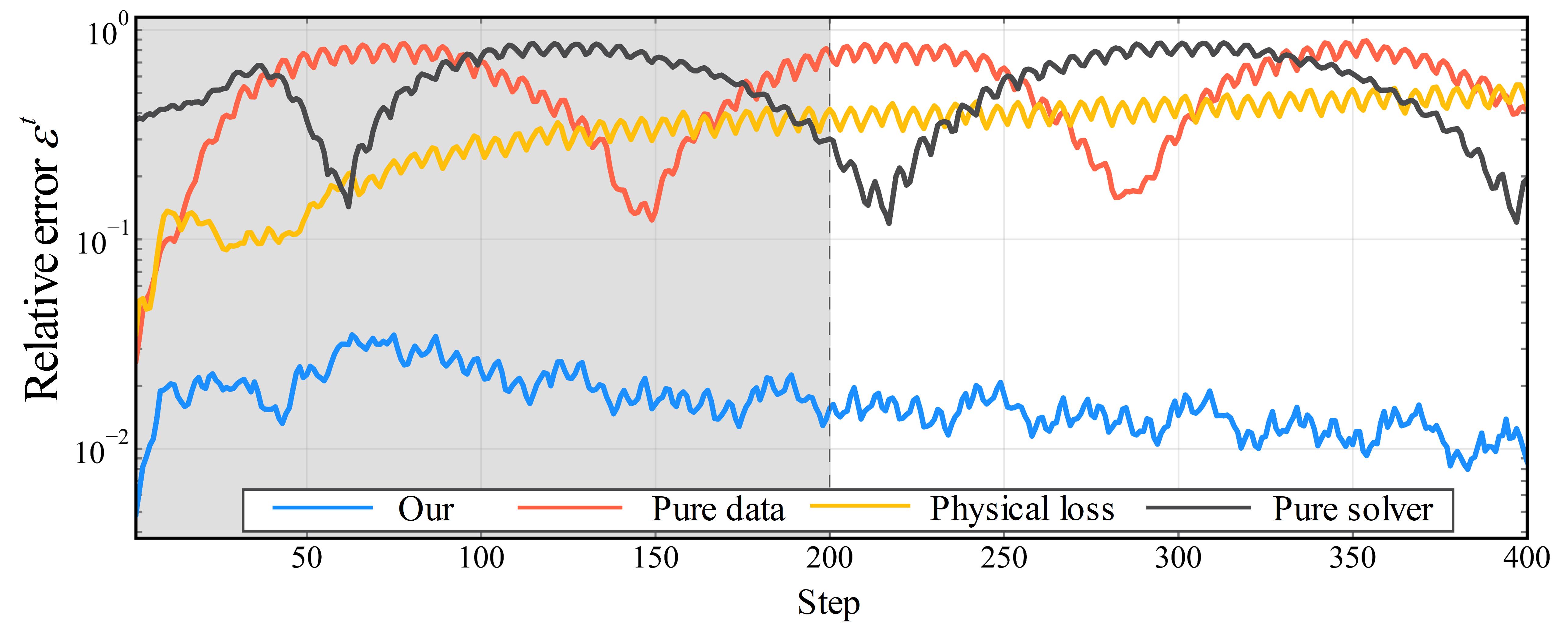}
\caption{Stepwise rollout relative error $\epsilon^{\,t}$ for flow past a rotationally oscillating circular cylinder at $Re=100$ and $\omega=4$, defined in Eq.~\eqref{eq:step_rel_error_t}. The metric is computed as the average of the relative $L_2$ errors of $u$, $v$, and the structural response vector $\mathbf{F}$ at each rollout step. The gray-shaded region indicates the training window, and the white region indicates temporal extrapolation beyond the training horizon.}
\label{r4_rerror}
\end{figure}

In contrast, the proposed physics-integrated neural differentiable model explicitly enforces the solid-boundary flow behavior through the IB method and computes the structural load response along a physically consistent update path. As a result, its advantage in this case is not merely higher instantaneous accuracy, but a stronger ability to maintain stable and physically admissible rollout trajectories in a more nonlinear and weakly periodic regime. This is particularly important because the non-lock-in condition rules out the possibility that good long-horizon performance is achieved simply by reproducing an approximately periodic pattern. Instead, the results indicate that the proposed model captures the underlying flow-evolution mechanism more faithfully. As evidenced by the error curves in Fig.~\ref{r4_rerror}, our model significantly outperforms the other baselines and exhibits almost no error accumulation under long-horizon rollouts, demonstrating superior stability in a genuinely challenging prediction scenario.

\section{Discussion} \label{5}
\subsection{Generalizability over unseen cases}

In the evaluations reported in Section 4, all models perform temporal extrapolation under seen operating conditions. To further assess generalization to unseen conditions a critical criterion for practical surrogate model deployment we consider the rotationally oscillating cylinder as an additional testbed. Specifically, the cases with rotation rates $\omega = 2$ and $\omega = 4$ are used for training, while $\omega = 3$ and $\omega = 5$ are reserved for testing, representing an interpolative and an extrapolative regime, respectively. The training data cover the nondimensional time window $t^* \in [0, 100]$, whereas for the test conditions the trained model is rolled out over $t^* \in [0, 200]$ to evaluate both in-horizon and out-of-horizon performance.

\begin{figure}[htbp]
\centering
\includegraphics[scale=.7]{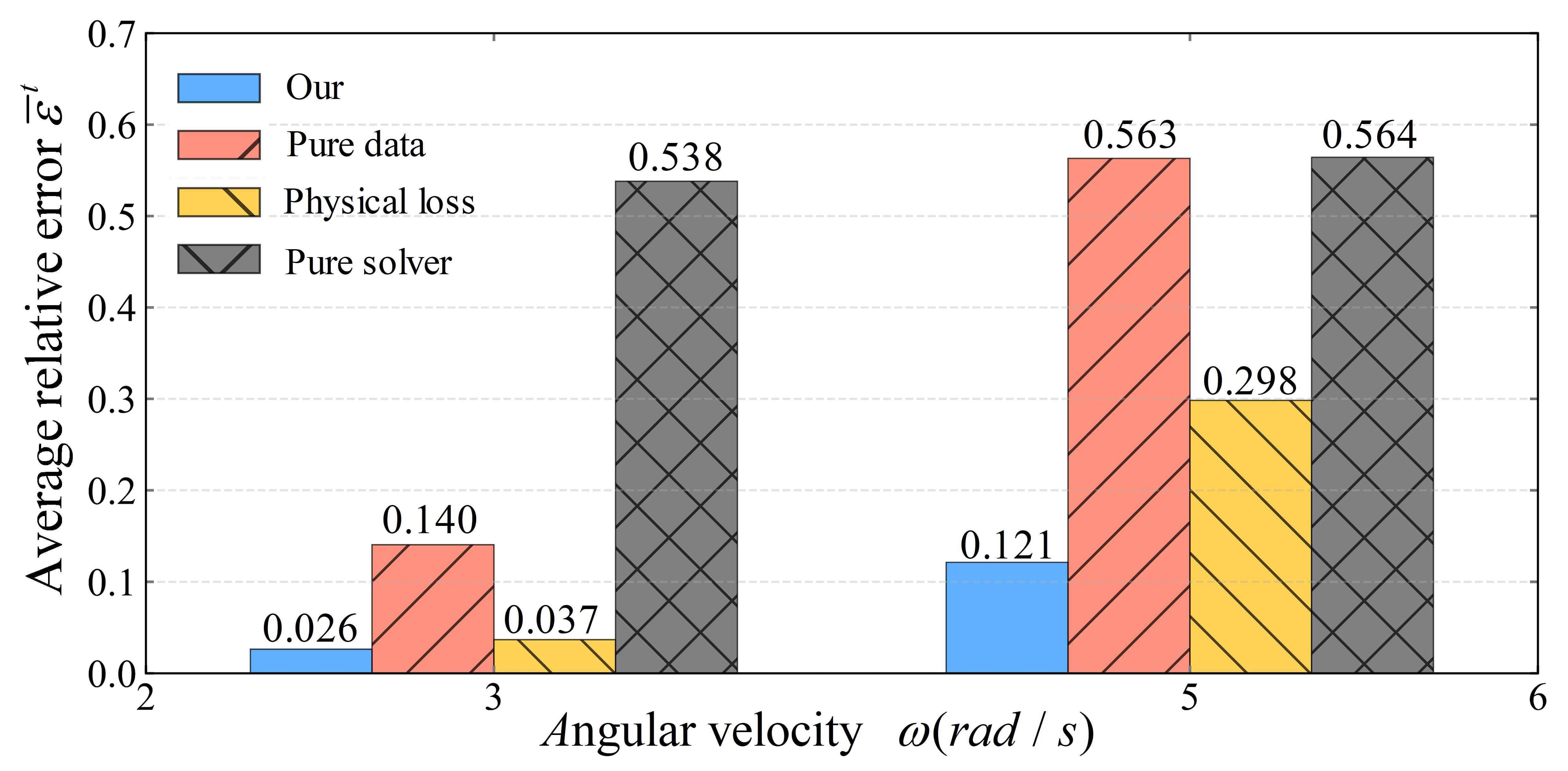}
\caption{Average relative error $\bar{\epsilon}$ (averaged over the entire inference horizon) under unseen operating conditions for the rotationally oscillating cylinder. Models are trained on $\omega=\{2,4\}$ and evaluated at $\omega=3$ (interpolation) and $\omega=5$ (extrapolation). Lower values indicate better long-horizon generalization.}
\label{avg_rerror}
\end{figure}

By examining the mean rollout error computed as the relative error averaged over the whole inference horizon the results in Fig.\ref{avg_rerror} show that, in the cross-case comparison, the proposed model achieves the best performance under both test conditions. In the within-model comparison, all methods perform better on the interpolative case than on the extrapolative case, which is consistent with the general expectation that interpolation is inherently easier than extrapolation. This is because the extrapolative condition is associated with a larger departure from the training regime, making the wake evolution and force response more difficult to reproduce accurately over long rollouts.

\begin{figure}[htbp]
\centering
\includegraphics[scale=.6]{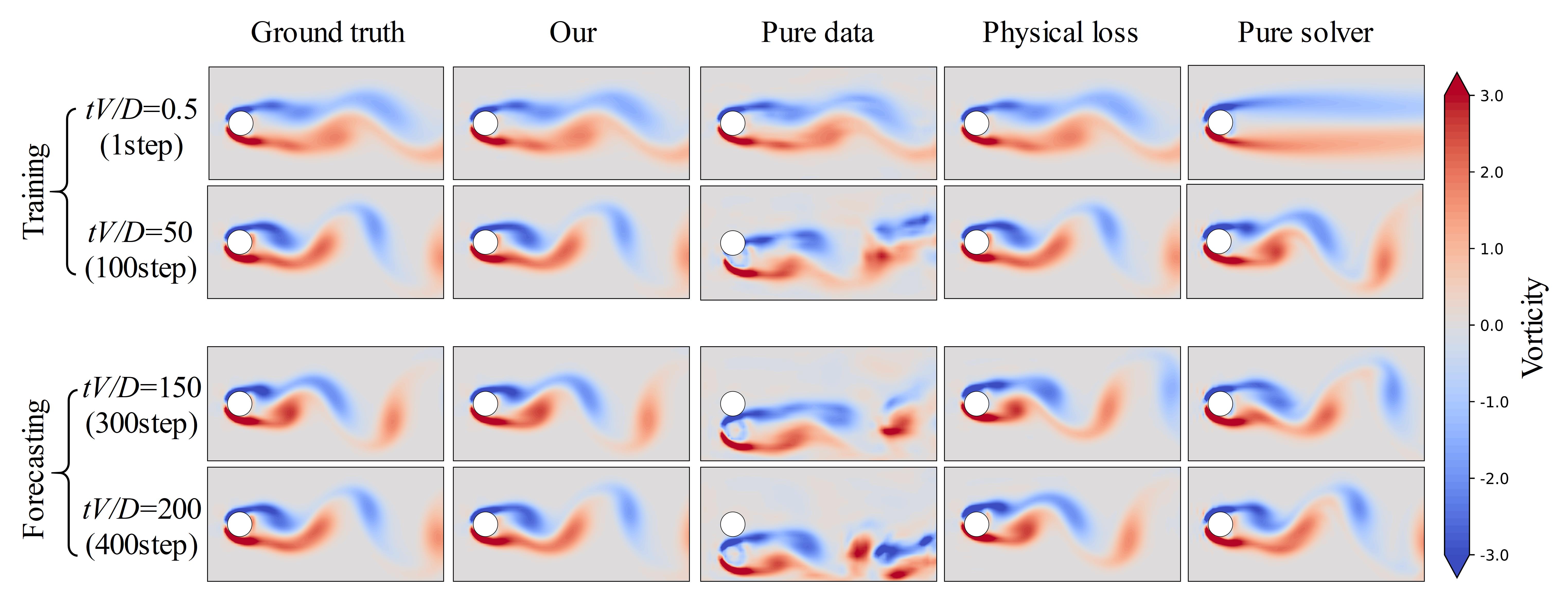}
\caption{Comparison of spatio-temporal vorticity predictions by different neural models for flow past a rotationally oscillating circular cylinder at $(Re=100, \omega=5)$.}
\label{r5_vortex}
\end{figure}

Taking the more challenging extrapolative case with $\omega=5$ as an example, we further analyze the model performance from both temporal and spatial perspectives. The results are shown in Figs.~\ref{r5_vortex} and \ref{force_r5}. The purely data-driven model and the physics-loss-constrained model exhibit behaviors largely consistent with the conclusions in Section 4, namely rapid error accumulation and eventual divergence. By contrast, the proposed model shows only a moderate degradation under extrapolation: as the rollout horizon increases, the errors in both the predicted flow field and the structural load response gradually accumulate, which is more clearly observed in Fig.~\ref{force_r5}. Nevertheless, benefiting from the constraints imposed by the physically structured architecture, the proposed model still delivers substantially more accurate rollouts than the other approaches, even under this more demanding unseen condition.

\begin{figure}[htbp]
\centering
\includegraphics[scale=.35]{figure/force_r5.jpg}
\caption{Comparison of structural force predictions by different neural models for flow past a rotationally oscillating circular cylinder at $(Re=100, \omega=5)$. The left column shows the time histories of the drag coefficient $C_D$, and the right column shows the lift coefficient $C_L$. The gray-shaded interval denotes the training window, whereas the white interval corresponds to out-of-distribution time extrapolation beyond the training horizon. The black solid line represents the ground truth, and the blue curve corresponds to the prediction of the proposed model.}
\label{force_r5}
\end{figure}

\subsection{Training cost and inference acceleration evaluation}

\begin{figure}[htbp]
\centering
\includegraphics[scale=.5]{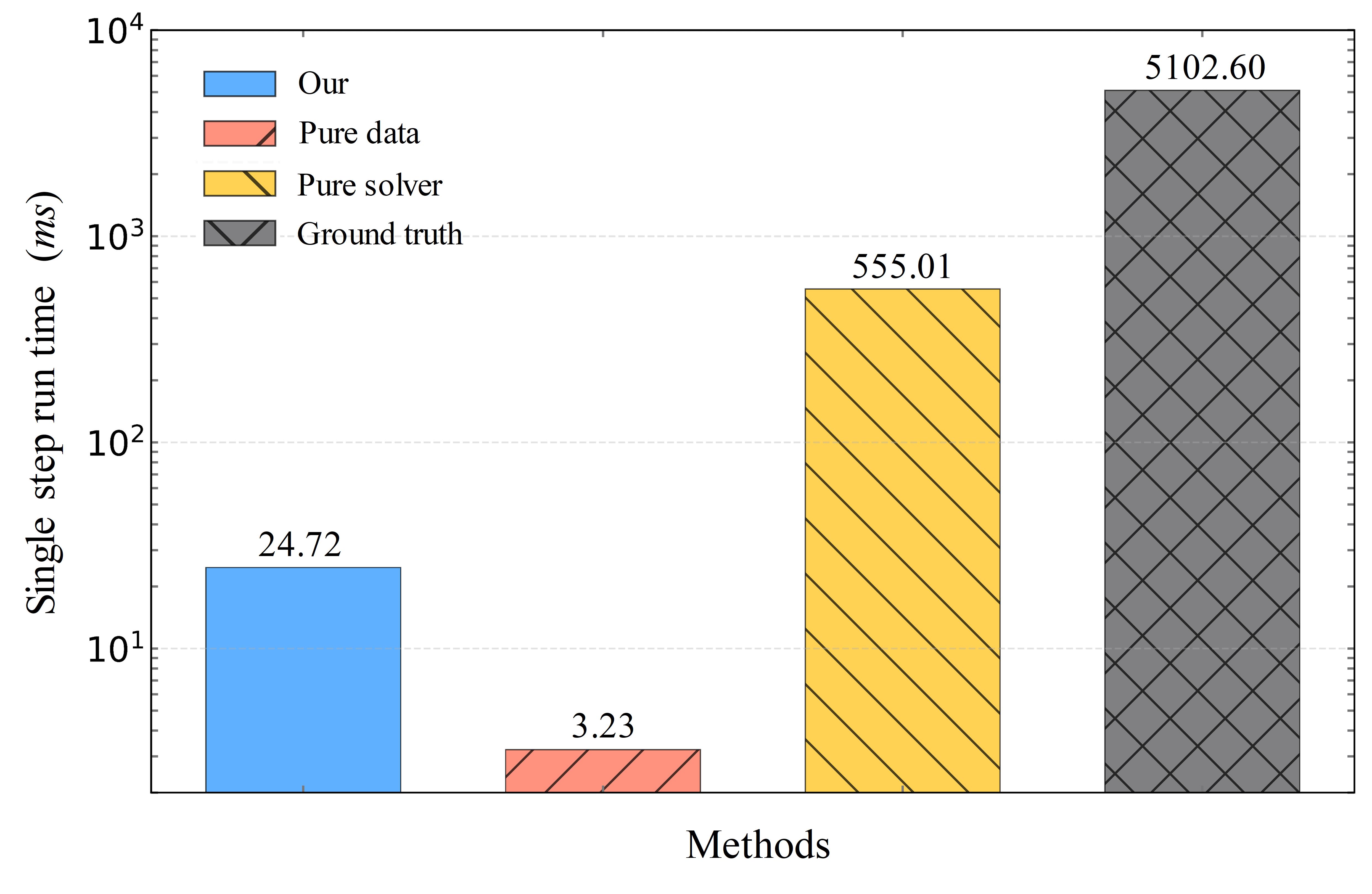}
\caption{Wall-clock time per learning step for different methods. The neural models are evaluated on the truncated subdomain $\Omega^{r}_{f}$ using an $80\times40$ grid, whereas the ground-truth CFD solution is computed on the full domain $\Omega_{f}$ using a $717\times382$ grid.}
\label{runtime}
\end{figure}

As shown in Fig.~\ref{runtime}, we report the wall-clock time required for a single learning step $\Delta t$ (corresponding to a time span of 100 numerical time steps) and compare it across models. The purely data-driven model achieves the shortest runtime, since it involves no explicit physics computation and relies solely on fast network inference, resulting in the lowest computational cost. The physics-loss-constrained model shares the same network architecture as the purely data-driven baseline and is therefore not shown separately. Owing to the coarse grid resolution and the larger time step size, the coarse numerical simulation is also substantially faster than the high-resolution solver, although it suffers from significant accuracy limitations.

The proposed model is slower than the purely data-driven baseline because it retains explicit physics-based computations within the inference process. However, compared with conventional numerical solvers, it permits a much larger effective time step—far beyond the CFL-limited step size required by the embedded physics solver itself—and avoids the expensive pressure-projection correction. As a result, on the same hardware platform (a single RTX 3080Ti GPU with 12 GB memory), the proposed approach achieves approximately a 200-fold inference speedup relative to the high-resolution numerical solver, and about a 20-fold speedup relative to the coarse-grid numerical solver at the same resolution, while maintaining good accuracy and physical consistency.

\begin{figure}[htbp]
\centering
\includegraphics[scale=.5]{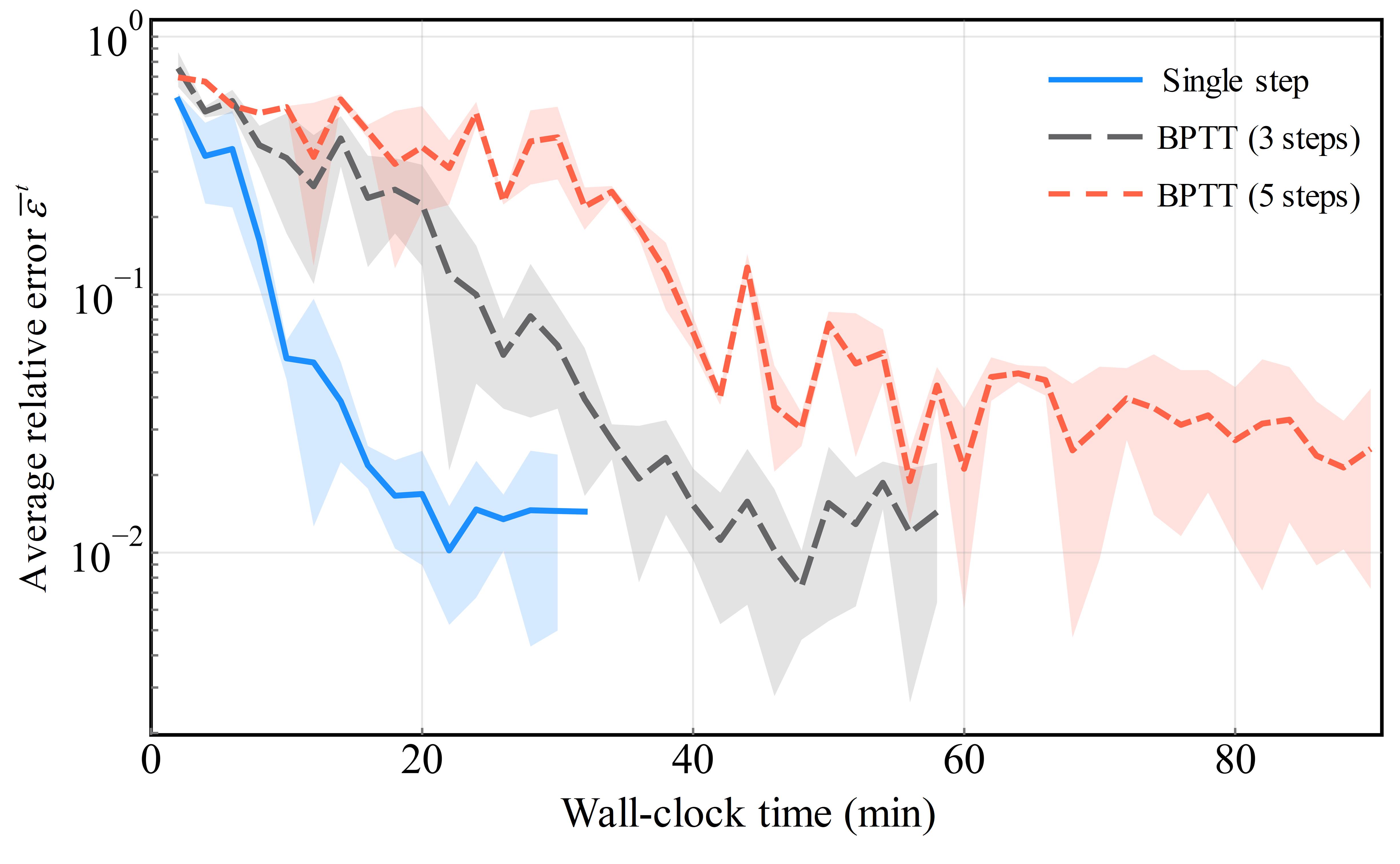}
\caption{Training efficiency comparison between the single-step and BPTT-based multi-step training for flow past a stationary circular cylinder at Reynolds number $Re=100$. The shaded bands represent $\pm \sigma$.}
\label{train_time}
\end{figure}

In terms of training cost, we further compare the single-step supervised strategy adopted in this work with backpropagation through time (BPTT) using temporal windows of length 3 and 5. All experiments are conducted on the same personal workstation equipped with a single RTX 3080Ti GPU (12 GB memory), using the same training hyperparameters to ensure a fair comparison. As shown in Fig.~\ref{train_time}, the single-step strategy and BPTT with a 3-step window achieve broadly comparable low-error performance, whereas BPTT with a 5-step window shows slower convergence and remains at a noticeably higher error level within the considered training time. This may be attributed to the limited model capacity and the finite amount of training data, which struggle to support the more complex optimization landscape of longer temporal windows. The competitive performance achieved with only single-step training is mainly enabled by the physically structured design of the proposed model, where the main numerical update path and key physical constraints are embedded directly into the forward propagation, thereby reducing the dependence on long unrolled supervision for learning stable temporal evolution. Consequently, within the present framework, the model can achieve competitive performance using only single-step training, without implying that BPTT is generally unnecessary for neural surrogate modeling. At the same time, the single-step setting is markedly more efficient with respect to wall-clock training time, because each training update only involves one-step forward prediction and gradient computation, whereas BPTT must unroll the model over a temporal window and backpropagate through the corresponding computational graph. Even for the relatively short windows considered here, this leads to substantially slower convergence in terms of wall-clock time.

A similar trend is observed in memory consumption. Because BPTT must retain intermediate activations over multiple rollout steps for gradient propagation, its memory usage increases significantly with the temporal window length, as summarized in Table~\ref{tab:memory_cost}. In practice, this rapidly growing memory demand is a major factor limiting the use of longer rollout horizons in training, and is one reason why many existing studies\cite{Wang_Chu_2025, xu2021accelerated} resort to truncated temporal windows to balance optimization quality against computational cost. By contrast, the proposed single-step training strategy avoids long-horizon graph construction altogether, enabling all cases in this work to be trained within one hour on a single consumer-grade GPU. This efficiency is particularly valuable in settings where training data must be generated from expensive numerical simulations and multiple rounds of model development are required.

\begin{table}[htbp]
\centering
\caption{GPU memory consumption under different training strategies.}
\label{tab:memory_cost}
\begin{tabular}{lc}
\toprule
Training strategy & GPU memory usage (GB) \\
\midrule
Single-step training & 2.7 \\
BPTT ($T=3$) & 5.7 \\
BPTT ($T=5$) & 8.8 \\
\bottomrule
\end{tabular}
\end{table}

\section{Conclusion} \label{6}

In this work, we have developed a physics-integrated neural differentiable surrogate for incompressible flows with immersed boundaries, targeting accurate, stable, and efficient long-horizon prediction. This work is an extension of the research by Fan and Wang, which further advances the field of neural surrogates for fluid–structure interaction. The primary contribution of this work lies in structurally integrating physical principles into the model architecture rather than treating physics as an auxiliary regularization term. This integration is a key distinction from existing physics-loss-constrained surrogates. Specifically, the proposed model embeds the pressure-projection update logic into an end-to-end differentiable framework, ensuring that the model architecture and data flow are consistent with incompressible flow computation and that each intermediate variable retains a clear physical meaning. Within this framework, we make three additional key contributions: an explicit immersed-boundary forcing module is incorporated to enforce solid-boundary constraints and enable physically consistent hydrodynamic force evaluation; a sub-iteration strategy decouples the embedded physics solver’s internal stability requirement from the surrogate’s time step to enable stable coarse-grid rollouts with large time steps, which overcomes a critical limitation of conventional coarse-grid solvers; and a learned correction module replaces the expensive pressure-Poisson projection to improve computational efficiency while preserving accuracy. Notably, the entire model can be trained via single-step supervision, which avoids long-horizon backpropagation and reduces training cost and memory consumption as another significant practical extension.

Extensive tests on canonical benchmarks, including flow past a stationary cylinder and a rotationally oscillating cylinder, validate our contributions. These tests show that the proposed model consistently outperforms purely data-driven surrogates, physics-loss-constrained surrogates, and coarse-grid numerical solvers in spatial fidelity and temporal robustness. The model accurately preserves near-body vortical structures, wake evolution, and structural load responses, while suppressing long-horizon error accumulation and phase drift. A further critical contribution is the validation of the model’s generalization to unseen operating conditions. Using interpolative and extrapolative rotation rates, the model achieves the lowest mean rollout error, even under the more challenging extrapolative regime, which demonstrates broader applicability beyond seen training conditions. More broadly, this work establishes a practical differentiable modeling framework for immersed-boundary flows. It extends the insights of Fan and Wang by demonstrating that physically structured architecture design, rather than loss-based regularization alone, provides a more effective route toward stable long-horizon prediction and efficient surrogate simulation.

These results confirm that the proposed approach is more than an accurate surrogate model. It delivers a novel, practical differentiable modeling framework for immersed-boundary flows, which is our overarching contribution and a significant extension of Fan and Wang with clear potential for scientific machine learning and differentiable simulation. By combining long-horizon stability, physical interpretability, and high computational efficiency in a single trainable architecture, the framework is well suited for applications requiring repeated flow evaluations, such as design optimization and flow control, as it addresses a key computational bottleneck. This advantage is particularly relevant for emerging data-driven and reinforcement-learning-based control paradigms, where the model can serve as an accurate and efficient environment surrogate. Owing to its fully differentiable formulation and modular design, the framework also provides a flexible foundation for future extensions to more complex fluid–structure interaction systems.

\section*{Author contribution}
Chenglin Li: Conceptualization, Methodology, Software, Formal analysis, 
Investigation, Writing -- original draft. 
Hang Xu: Conceptualization, Resources, Supervision, Writing -- review \& editing.
Jianting Chen: Draft revision.
Yanfei Zhang: Draft revision.

\section*{Data availability}
The data that support the findings of this study are available from the corresponding author upon request. 

\section*{Acknowledgment}
This work was supported by the open fund project of the State Key Laboratory of Maritime Technology and Safety (Grant No.W25CG000074).

\bibliographystyle{elsarticle-num-names}

\bibliography{cas-refs}

\begin{thebibliography}{49}
\expandafter\ifx\csname natexlab\endcsname\relax\def\natexlab#1{#1}\fi
\providecommand{\url}[1]{\texttt{#1}}
\providecommand{\href}[2]{#2}
\providecommand{\path}[1]{#1}
\providecommand{\DOIprefix}{doi:}
\providecommand{\ArXivprefix}{arXiv:}
\providecommand{\URLprefix}{URL: }
\providecommand{\Pubmedprefix}{pmid:}
\providecommand{\doi}[1]{\href{http://dx.doi.org/#1}{\path{#1}}}
\providecommand{\Pubmed}[1]{\href{pmid:#1}{\path{#1}}}
\providecommand{\bibinfo}[2]{#2}
\ifx\xfnm\relax \def\xfnm[#1]{\unskip,\space#1}\fi
\bibitem[{Rabault et~al.(2019)Rabault, Kuchta, Jensen, Réglade, and Cerardi}]{Rabault_2019}
\bibinfo{author}{J.~Rabault}, \bibinfo{author}{M.~Kuchta}, \bibinfo{author}{A.~Jensen}, \bibinfo{author}{U.~Réglade}, \bibinfo{author}{N.~Cerardi},
\newblock \bibinfo{title}{Artificial neural networks trained through deep reinforcement learning discover control strategies for active flow control},
\newblock \bibinfo{journal}{Journal of Fluid Mechanics} \bibinfo{volume}{865} (\bibinfo{year}{2019}) \bibinfo{pages}{281–302}. \DOIprefix\doi{https://doi.org/10.1017/jfm.2019.62}.
\bibitem[{Costanzo et~al.(2022)Costanzo, Sayadi, {Fosas de Pando}, Schmid, and Frey}]{COSTANZO2022111664}
\bibinfo{author}{S.~Costanzo}, \bibinfo{author}{T.~Sayadi}, \bibinfo{author}{M.~{Fosas de Pando}}, \bibinfo{author}{P.~Schmid}, \bibinfo{author}{P.~Frey},
\newblock \bibinfo{title}{Parallel-in-time adjoint-based optimization – application to unsteady incompressible flows},
\newblock \bibinfo{journal}{Journal of Computational Physics} \bibinfo{volume}{471} (\bibinfo{year}{2022}) \bibinfo{pages}{111664}. \DOIprefix\doi{https://doi.org/10.1016/j.jcp.2022.111664}.
\bibitem[{Jia and Xu(2025)}]{JIA2025111125}
\bibinfo{author}{W.~Jia}, \bibinfo{author}{H.~Xu},
\newblock \bibinfo{title}{Strategies for energy-efficient flow control leveraging deep reinforcement learning},
\newblock \bibinfo{journal}{Engineering Applications of Artificial Intelligence} \bibinfo{volume}{156} (\bibinfo{year}{2025}) \bibinfo{pages}{111125}. \DOIprefix\doi{https://doi.org/10.1016/j.engappai.2025.111125}.
\bibitem[{Jia and Xu(2024)}]{10.1063/5.0204237}
\bibinfo{author}{W.~Jia}, \bibinfo{author}{H.~Xu},
\newblock \bibinfo{title}{Optimal parallelization strategies for active flow control in deep reinforcement learning-based computational fluid dynamics},
\newblock \bibinfo{journal}{Physics of Fluids} \bibinfo{volume}{36} (\bibinfo{year}{2024}) \bibinfo{pages}{043623}. \DOIprefix\doi{10.1063/5.0204237}.
\bibitem[{Peskin(2002)}]{Peskin_2002}
\bibinfo{author}{C.~S. Peskin},
\newblock \bibinfo{title}{The immersed boundary method},
\newblock \bibinfo{journal}{Acta Numerica} \bibinfo{volume}{11} (\bibinfo{year}{2002}) \bibinfo{pages}{479–517}. \DOIprefix\doi{10.1017/S0962492902000077}.
\bibitem[{Thompson et~al.(1982)Thompson, Warsi, and {Wayne Mastin}}]{THOMPSON19821}
\bibinfo{author}{J.~F. Thompson}, \bibinfo{author}{Z.~U. Warsi}, \bibinfo{author}{C.~{Wayne Mastin}},
\newblock \bibinfo{title}{Boundary-fitted coordinate systems for numerical solution of partial differential equations—a review},
\newblock \bibinfo{journal}{Journal of Computational Physics} \bibinfo{volume}{47} (\bibinfo{year}{1982}) \bibinfo{pages}{1--108}. \DOIprefix\doi{https://doi.org/10.1016/0021-9991(82)90066-3}.
\bibitem[{Johnson and Tezduyar(1994)}]{JOHNSON199473}
\bibinfo{author}{A.~Johnson}, \bibinfo{author}{T.~Tezduyar},
\newblock \bibinfo{title}{Mesh update strategies in parallel finite element computations of flow problems with moving boundaries and interfaces},
\newblock \bibinfo{journal}{Computer Methods in Applied Mechanics and Engineering} \bibinfo{volume}{119} (\bibinfo{year}{1994}) \bibinfo{pages}{73--94}. \DOIprefix\doi{https://doi.org/10.1016/0045-7825(94)00077-8}.
\bibitem[{Mittal and Iaccarino(2005)}]{annurev.fluid.37.061903.175743}
\bibinfo{author}{R.~Mittal}, \bibinfo{author}{G.~Iaccarino},
\newblock \bibinfo{title}{Immersed boundary methods},
\newblock \bibinfo{journal}{Annual Review of Fluid Mechanics} \bibinfo{volume}{37} (\bibinfo{year}{2005}) \bibinfo{pages}{239--261}. \DOIprefix\doi{https://doi.org/10.1146/annurev.fluid.37.061903.175743}.
\bibitem[{Yang and Stern(2012)}]{YANG20125029}
\bibinfo{author}{J.~Yang}, \bibinfo{author}{F.~Stern},
\newblock \bibinfo{title}{A simple and efficient direct forcing immersed boundary framework for fluid–structure interactions},
\newblock \bibinfo{journal}{Journal of Computational Physics} \bibinfo{volume}{231} (\bibinfo{year}{2012}) \bibinfo{pages}{5029--5061}. \DOIprefix\doi{https://doi.org/10.1016/j.jcp.2012.04.012}.
\bibitem[{Zhu et~al.(2026)Zhu, Yang, Du, and Yu}]{ZHU2026106913}
\bibinfo{author}{H.~Zhu}, \bibinfo{author}{Y.~Yang}, \bibinfo{author}{Z.~Du}, \bibinfo{author}{J.~Yu},
\newblock \bibinfo{title}{Gpu accelerated vortex-induced vibration simulation using jax: Efficiency and accuracy strategies},
\newblock \bibinfo{journal}{Computers \& Fluids} \bibinfo{volume}{305} (\bibinfo{year}{2026}) \bibinfo{pages}{106913}. \DOIprefix\doi{https://doi.org/10.1016/j.compfluid.2025.106913}.
\bibitem[{Giannenas and Laizet(2021)}]{GIANNENAS2021606}
\bibinfo{author}{A.~E. Giannenas}, \bibinfo{author}{S.~Laizet},
\newblock \bibinfo{title}{A simple and scalable immersed boundary method for high-fidelity simulations of fixed and moving objects on a cartesian mesh},
\newblock \bibinfo{journal}{Applied Mathematical Modelling} \bibinfo{volume}{99} (\bibinfo{year}{2021}) \bibinfo{pages}{606--627}. \DOIprefix\doi{https://doi.org/10.1016/j.apm.2021.06.026}.
\bibitem[{Vinuesa and Brunton(2022)}]{Vinuesa2022}
\bibinfo{author}{R.~Vinuesa}, \bibinfo{author}{S.~L. Brunton},
\newblock \bibinfo{title}{Enhancing computational fluid dynamics with machine learning},
\newblock \bibinfo{journal}{Nature Computational Science} \bibinfo{volume}{2} (\bibinfo{year}{2022}) \bibinfo{pages}{358--366}. \DOIprefix\doi{10.1038/s43588-022-00264-7}.
\bibitem[{Davydzenka and Tahmasebi(2022)}]{Davydzenka_Tahmasebi_2022}
\bibinfo{author}{T.~Davydzenka}, \bibinfo{author}{P.~Tahmasebi},
\newblock \bibinfo{title}{High-resolution fluid–particle interactions: a machine learning approach},
\newblock \bibinfo{journal}{Journal of Fluid Mechanics} \bibinfo{volume}{938} (\bibinfo{year}{2022}) \bibinfo{pages}{A20}. \DOIprefix\doi{10.1017/jfm.2022.174}.
\bibitem[{Taira et~al.(2017)Taira, Brunton, Dawson, Rowley, Colonius, McKeon, Schmidt, Gordeyev, Theofilis, and Ukeiley}]{doi:10.2514/1.J056060}
\bibinfo{author}{K.~Taira}, \bibinfo{author}{S.~L. Brunton}, \bibinfo{author}{S.~T.~M. Dawson}, \bibinfo{author}{C.~W. Rowley}, \bibinfo{author}{T.~Colonius}, \bibinfo{author}{B.~J. McKeon}, \bibinfo{author}{O.~T. Schmidt}, \bibinfo{author}{S.~Gordeyev}, \bibinfo{author}{V.~Theofilis}, \bibinfo{author}{L.~S. Ukeiley},
\newblock \bibinfo{title}{Modal analysis of fluid flows: An overview},
\newblock \bibinfo{journal}{AIAA Journal} \bibinfo{volume}{55} (\bibinfo{year}{2017}) \bibinfo{pages}{4013--4041}. \DOIprefix\doi{10.2514/1.J056060}.
\bibitem[{Murata et~al.(2020)Murata, Fukami, and Fukagata}]{Murata_Fukami_Fukagata_2020}
\bibinfo{author}{T.~Murata}, \bibinfo{author}{K.~Fukami}, \bibinfo{author}{K.~Fukagata},
\newblock \bibinfo{title}{Nonlinear mode decomposition with convolutional neural networks for fluid dynamics},
\newblock \bibinfo{journal}{Journal of Fluid Mechanics} \bibinfo{volume}{882} (\bibinfo{year}{2020}) \bibinfo{pages}{A13}. \DOIprefix\doi{10.1017/jfm.2019.822}.
\bibitem[{Cremades et~al.(2024)Cremades, Hoyas, Deshpande, Quintero, Lellep, Lee, Monty, Hutchins, Linkmann, Marusic, and Vinuesa}]{Cremades2024}
\bibinfo{author}{A.~Cremades}, \bibinfo{author}{S.~Hoyas}, \bibinfo{author}{R.~Deshpande}, \bibinfo{author}{P.~Quintero}, \bibinfo{author}{M.~Lellep}, \bibinfo{author}{W.~J. Lee}, \bibinfo{author}{J.~P. Monty}, \bibinfo{author}{N.~Hutchins}, \bibinfo{author}{M.~Linkmann}, \bibinfo{author}{I.~Marusic}, \bibinfo{author}{R.~Vinuesa},
\newblock \bibinfo{title}{Identifying regions of importance in wall-bounded turbulence through explainable deep learning},
\newblock \bibinfo{journal}{Nature Communications} \bibinfo{volume}{15} (\bibinfo{year}{2024}) \bibinfo{pages}{3864}. \DOIprefix\doi{10.1038/s41467-024-47954-6}.
\bibitem[{Li et~al.(2023)Li, Buzzicotti, Biferale, Bonaccorso, Chen, and Wan}]{Li_Buzzicotti_2023}
\bibinfo{author}{T.~Li}, \bibinfo{author}{M.~Buzzicotti}, \bibinfo{author}{L.~Biferale}, \bibinfo{author}{F.~Bonaccorso}, \bibinfo{author}{S.~Chen}, \bibinfo{author}{M.~Wan},
\newblock \bibinfo{title}{Multi-scale reconstruction of turbulent rotating flows with proper orthogonal decomposition and generative adversarial networks},
\newblock \bibinfo{journal}{Journal of Fluid Mechanics} \bibinfo{volume}{971} (\bibinfo{year}{2023}) \bibinfo{pages}{A3}. \DOIprefix\doi{10.1017/jfm.2023.573}.
\bibitem[{Hadizadeh et~al.(2025)Hadizadeh, Mallik, and Jaiman}]{HADIZADEH2025117921}
\bibinfo{author}{F.~Hadizadeh}, \bibinfo{author}{W.~Mallik}, \bibinfo{author}{R.~K. Jaiman},
\newblock \bibinfo{title}{A graph neural network surrogate model for multi-objective fluid-acoustic shape optimization},
\newblock \bibinfo{journal}{Computer Methods in Applied Mechanics and Engineering} \bibinfo{volume}{441} (\bibinfo{year}{2025}) \bibinfo{pages}{117921}. \DOIprefix\doi{https://doi.org/10.1016/j.cma.2025.117921}.
\bibitem[{Barwey et~al.(2025)Barwey, Kim, and Maulik}]{BARWEY2025117509}
\bibinfo{author}{S.~Barwey}, \bibinfo{author}{H.~Kim}, \bibinfo{author}{R.~Maulik},
\newblock \bibinfo{title}{Interpretable a-posteriori error indication for graph neural network surrogate models},
\newblock \bibinfo{journal}{Computer Methods in Applied Mechanics and Engineering} \bibinfo{volume}{433} (\bibinfo{year}{2025}) \bibinfo{pages}{117509}. \DOIprefix\doi{https://doi.org/10.1016/j.cma.2024.117509}.
\bibitem[{Brunton et~al.(2020)Brunton, Noack, and Koumoutsakos}]{annurev-fluid-010719-060214}
\bibinfo{author}{S.~L. Brunton}, \bibinfo{author}{B.~R. Noack}, \bibinfo{author}{P.~Koumoutsakos},
\newblock \bibinfo{title}{Machine learning for fluid mechanics},
\newblock \bibinfo{journal}{Annual Review of Fluid Mechanics} \bibinfo{volume}{52} (\bibinfo{year}{2020}) \bibinfo{pages}{477--508}. \DOIprefix\doi{https://doi.org/10.1146/annurev-fluid-010719-060214}.
\bibitem[{Liang et~al.(2022)Liang, Tadesse, Ho, Fei-Fei, Zaharia, Zhang, and Zou}]{Liang2022}
\bibinfo{author}{W.~Liang}, \bibinfo{author}{G.~A. Tadesse}, \bibinfo{author}{D.~Ho}, \bibinfo{author}{L.~Fei-Fei}, \bibinfo{author}{M.~Zaharia}, \bibinfo{author}{C.~Zhang}, \bibinfo{author}{J.~Zou},
\newblock \bibinfo{title}{Advances, challenges and opportunities in creating data for trustworthy ai},
\newblock \bibinfo{journal}{Nature Machine Intelligence} \bibinfo{volume}{4} (\bibinfo{year}{2022}) \bibinfo{pages}{669--677}. \DOIprefix\doi{10.1038/s42256-022-00516-1}.
\bibitem[{Karniadakis et~al.(2021)Karniadakis, Kevrekidis, Lu, Perdikaris, Wang, and Yang}]{Karniadakis2021}
\bibinfo{author}{G.~E. Karniadakis}, \bibinfo{author}{I.~G. Kevrekidis}, \bibinfo{author}{L.~Lu}, \bibinfo{author}{P.~Perdikaris}, \bibinfo{author}{S.~Wang}, \bibinfo{author}{L.~Yang},
\newblock \bibinfo{title}{Physics-informed machine learning},
\newblock \bibinfo{journal}{Nature Reviews Physics} \bibinfo{volume}{3} (\bibinfo{year}{2021}) \bibinfo{pages}{422--440}. \DOIprefix\doi{10.1038/s42254-021-00314-5}.
\bibitem[{Tompson et~al.(2017)Tompson, Schlachter, Sprechmann, and Perlin}]{10.5555/3305890.3306035}
\bibinfo{author}{J.~Tompson}, \bibinfo{author}{K.~Schlachter}, \bibinfo{author}{P.~Sprechmann}, \bibinfo{author}{K.~Perlin},
\newblock \bibinfo{title}{Accelerating eulerian fluid simulation with convolutional networks},
\newblock in: \bibinfo{booktitle}{Proceedings of the 34th International Conference on Machine Learning - Volume 70}, ICML'17, \bibinfo{publisher}{JMLR.org}, \bibinfo{year}{2017}, p. \bibinfo{pages}{3424–3433}.
\bibitem[{Kochkov et~al.(2021)Kochkov, Smith, Alieva, Wang, Brenner, and Hoyer}]{doi:10.1073/pnas.2101784118}
\bibinfo{author}{D.~Kochkov}, \bibinfo{author}{J.~A. Smith}, \bibinfo{author}{A.~Alieva}, \bibinfo{author}{Q.~Wang}, \bibinfo{author}{M.~P. Brenner}, \bibinfo{author}{S.~Hoyer},
\newblock \bibinfo{title}{Machine learning–accelerated computational fluid dynamics},
\newblock \bibinfo{journal}{Proceedings of the National Academy of Sciences} \bibinfo{volume}{118} (\bibinfo{year}{2021}) \bibinfo{pages}{e2101784118}. \DOIprefix\doi{10.1073/pnas.2101784118}.
\bibitem[{Liu et~al.(2024)Liu, Zhu, Lu, Sun, and Wang}]{Liu2024}
\bibinfo{author}{X.-Y. Liu}, \bibinfo{author}{M.~Zhu}, \bibinfo{author}{L.~Lu}, \bibinfo{author}{H.~Sun}, \bibinfo{author}{J.-X. Wang},
\newblock \bibinfo{title}{Multi-resolution partial differential equations preserved learning framework for spatiotemporal dynamics},
\newblock \bibinfo{journal}{Communications Physics} \bibinfo{volume}{7} (\bibinfo{year}{2024}) \bibinfo{pages}{31}. \DOIprefix\doi{10.1038/s42005-024-01521-z}.
\bibitem[{Jin et~al.(2021)Jin, Cai, Li, and Karniadakis}]{JIN2021109951}
\bibinfo{author}{X.~Jin}, \bibinfo{author}{S.~Cai}, \bibinfo{author}{H.~Li}, \bibinfo{author}{G.~E. Karniadakis},
\newblock \bibinfo{title}{Nsfnets (navier-stokes flow nets): Physics-informed neural networks for the incompressible navier-stokes equations},
\newblock \bibinfo{journal}{Journal of Computational Physics} \bibinfo{volume}{426} (\bibinfo{year}{2021}) \bibinfo{pages}{109951}. \DOIprefix\doi{https://doi.org/10.1016/j.jcp.2020.109951}.
\bibitem[{Zhu et~al.(2024)Zhu, Jiang, Lefauve, Kerswell, and Linden}]{Zhu_Jiang_2024}
\bibinfo{author}{L.~Zhu}, \bibinfo{author}{X.~Jiang}, \bibinfo{author}{A.~Lefauve}, \bibinfo{author}{R.~R. Kerswell}, \bibinfo{author}{P.~Linden},
\newblock \bibinfo{title}{New insights into experimental stratified flows obtained through physics-informed neural networks},
\newblock \bibinfo{journal}{Journal of Fluid Mechanics} \bibinfo{volume}{981} (\bibinfo{year}{2024}) \bibinfo{pages}{R1}. \DOIprefix\doi{10.1017/jfm.2024.49}.
\bibitem[{Haghighat et~al.(2021)Haghighat, Raissi, Moure, Gomez, and Juanes}]{HAGHIGHAT2021113741}
\bibinfo{author}{E.~Haghighat}, \bibinfo{author}{M.~Raissi}, \bibinfo{author}{A.~Moure}, \bibinfo{author}{H.~Gomez}, \bibinfo{author}{R.~Juanes},
\newblock \bibinfo{title}{A physics-informed deep learning framework for inversion and surrogate modeling in solid mechanics},
\newblock \bibinfo{journal}{Computer Methods in Applied Mechanics and Engineering} \bibinfo{volume}{379} (\bibinfo{year}{2021}) \bibinfo{pages}{113741}. \DOIprefix\doi{https://doi.org/10.1016/j.cma.2021.113741}.
\bibitem[{Cai et~al.(2021)Cai, Wang, Wang, Perdikaris, and Karniadakis}]{10.1115/1.4050542}
\bibinfo{author}{S.~Cai}, \bibinfo{author}{Z.~Wang}, \bibinfo{author}{S.~Wang}, \bibinfo{author}{P.~Perdikaris}, \bibinfo{author}{G.~E. Karniadakis},
\newblock \bibinfo{title}{Physics-informed neural networks for heat transfer problems},
\newblock \bibinfo{journal}{Journal of Heat Transfer} \bibinfo{volume}{143} (\bibinfo{year}{2021}) \bibinfo{pages}{060801}. \DOIprefix\doi{10.1115/1.4050542}.
\bibitem[{Baydin et~al.(2018)Baydin, Pearlmutter, Radul, and Siskind}]{JMLR:v18:17-468}
\bibinfo{author}{A.~G. Baydin}, \bibinfo{author}{B.~A. Pearlmutter}, \bibinfo{author}{A.~A. Radul}, \bibinfo{author}{J.~M. Siskind},
\newblock \bibinfo{title}{Automatic differentiation in machine learning: a survey},
\newblock \bibinfo{journal}{Journal of Machine Learning Research} \bibinfo{volume}{18} (\bibinfo{year}{2018}) \bibinfo{pages}{1--43}.
\bibitem[{Krishnapriyan et~al.(2021)Krishnapriyan, Gholami, Zhe, Kirby, and Mahoney}]{NEURIPS2021_df438e52}
\bibinfo{author}{A.~Krishnapriyan}, \bibinfo{author}{A.~Gholami}, \bibinfo{author}{S.~Zhe}, \bibinfo{author}{R.~Kirby}, \bibinfo{author}{M.~W. Mahoney},
\newblock \bibinfo{title}{Characterizing possible failure modes in physics-informed neural networks},
\newblock in: \bibinfo{editor}{M.~Ranzato}, \bibinfo{editor}{A.~Beygelzimer}, \bibinfo{editor}{Y.~Dauphin}, \bibinfo{editor}{P.~Liang}, \bibinfo{editor}{J.~W. Vaughan} (Eds.), \bibinfo{booktitle}{Advances in Neural Information Processing Systems}, volume~\bibinfo{volume}{34}, \bibinfo{publisher}{Curran Associates, Inc.}, \bibinfo{year}{2021}, pp. \bibinfo{pages}{26548--26560}.
\bibitem[{Chen et~al.(2018)Chen, Rubanova, Bettencourt, and Duvenaud}]{NEURIPS2018_69386f6b}
\bibinfo{author}{R.~T.~Q. Chen}, \bibinfo{author}{Y.~Rubanova}, \bibinfo{author}{J.~Bettencourt}, \bibinfo{author}{D.~Duvenaud},
\newblock \bibinfo{title}{Neural ordinary differential equations},
\newblock in: \bibinfo{booktitle}{Proceedings of the 32nd International Conference on Neural Information Processing Systems}, NIPS'18, \bibinfo{publisher}{Curran Associates Inc.}, \bibinfo{address}{Red Hook, NY, USA}, \bibinfo{year}{2018}, p. \bibinfo{pages}{6572–6583}.
\bibitem[{Long et~al.(2018)Long, Lu, Ma, and Dong}]{pmlr-v80-long18a}
\bibinfo{author}{Z.~Long}, \bibinfo{author}{Y.~Lu}, \bibinfo{author}{X.~Ma}, \bibinfo{author}{B.~Dong},
\newblock \bibinfo{title}{{PDE}-net: Learning {PDE}s from data},
\newblock in: \bibinfo{editor}{J.~Dy}, \bibinfo{editor}{A.~Krause} (Eds.), \bibinfo{booktitle}{Proceedings of the 35th International Conference on Machine Learning}, volume~\bibinfo{volume}{80} of \textit{\bibinfo{series}{Proceedings of Machine Learning Research}}, \bibinfo{publisher}{PMLR}, \bibinfo{year}{2018}, pp. \bibinfo{pages}{3208--3216}.
\bibitem[{Fan et~al.(2025)Fan, Liu, Wang, and Wang}]{fan2025diffflowf}
\bibinfo{author}{X.~Fan}, \bibinfo{author}{X.~Liu}, \bibinfo{author}{M.~Wang}, \bibinfo{author}{J.-X. Wang}, \bibinfo{title}{Diff-flowfsi: A gpu-optimized differentiable cfd platform for high-fidelity turbulence and fsi simulations}, \bibinfo{year}{2025}. \href{http://arxiv.org/abs/2505.23940}{{\tt arXiv:2505.23940}}.
\bibitem[{Bezgin et~al.(2023)Bezgin, Buhendwa, and Adams}]{BEZGIN2023108527}
\bibinfo{author}{D.~A. Bezgin}, \bibinfo{author}{A.~B. Buhendwa}, \bibinfo{author}{N.~A. Adams},
\newblock \bibinfo{title}{Jax-fluids: A fully-differentiable high-order computational fluid dynamics solver for compressible two-phase flows},
\newblock \bibinfo{journal}{Computer Physics Communications} \bibinfo{volume}{282} (\bibinfo{year}{2023}) \bibinfo{pages}{108527}. \DOIprefix\doi{https://doi.org/10.1016/j.cpc.2022.108527}.
\bibitem[{Fan and Wang(2024)}]{FAN2024112584}
\bibinfo{author}{X.~Fan}, \bibinfo{author}{J.-X. Wang},
\newblock \bibinfo{title}{Differentiable hybrid neural modeling for fluid-structure interaction},
\newblock \bibinfo{journal}{Journal of Computational Physics} \bibinfo{volume}{496} (\bibinfo{year}{2024}) \bibinfo{pages}{112584}. \DOIprefix\doi{https://doi.org/10.1016/j.jcp.2023.112584}.
\bibitem[{Fan et~al.(2025)Fan, Akhare, and Wang}]{FAN2025117478}
\bibinfo{author}{X.~Fan}, \bibinfo{author}{D.~Akhare}, \bibinfo{author}{J.-X. Wang},
\newblock \bibinfo{title}{Neural differentiable modeling with diffusion-based super-resolution for two-dimensional spatiotemporal turbulence},
\newblock \bibinfo{journal}{Computer Methods in Applied Mechanics and Engineering} \bibinfo{volume}{433} (\bibinfo{year}{2025}) \bibinfo{pages}{117478}. \DOIprefix\doi{https://doi.org/10.1016/j.cma.2024.117478}.
\bibitem[{Akhare et~al.(2023)Akhare, Luo, and Wang}]{AKHARE2023115902}
\bibinfo{author}{D.~Akhare}, \bibinfo{author}{T.~Luo}, \bibinfo{author}{J.-X. Wang},
\newblock \bibinfo{title}{Physics-integrated neural differentiable (pindiff) model for composites manufacturing},
\newblock \bibinfo{journal}{Computer Methods in Applied Mechanics and Engineering} \bibinfo{volume}{406} (\bibinfo{year}{2023}) \bibinfo{pages}{115902}. \DOIprefix\doi{https://doi.org/10.1016/j.cma.2023.115902}.
\bibitem[{Wang and Chu(2025)}]{Wang_Chu_2025}
\bibinfo{author}{W.~Wang}, \bibinfo{author}{X.~Chu},
\newblock \bibinfo{title}{Optimised flow control based on automatic differentiation in compressible turbulent channel flows},
\newblock \bibinfo{journal}{Journal of Fluid Mechanics} \bibinfo{volume}{1011} (\bibinfo{year}{2025}) \bibinfo{pages}{A1}. \DOIprefix\doi{10.1017/jfm.2025.304}.
\bibitem[{Xu et~al.(2021)Xu, Makoviychuk, Narang, Ramos, Matusik, Garg, and Macklin}]{xu2021accelerated}
\bibinfo{author}{J.~Xu}, \bibinfo{author}{V.~Makoviychuk}, \bibinfo{author}{Y.~Narang}, \bibinfo{author}{F.~Ramos}, \bibinfo{author}{W.~Matusik}, \bibinfo{author}{A.~Garg}, \bibinfo{author}{M.~Macklin},
\newblock \bibinfo{title}{Accelerated policy learning with parallel differentiable simulation},
\newblock in: \bibinfo{booktitle}{International Conference on Learning Representations}, \bibinfo{year}{2021}, pp. \bibinfo{pages}{1--26}.
\bibitem[{Kumar et~al.(2013)Kumar, Lopez, Probst, Francisco, Askari, and Yang}]{Kumar_Lopez2013}
\bibinfo{author}{S.~Kumar}, \bibinfo{author}{C.~Lopez}, \bibinfo{author}{O.~Probst}, \bibinfo{author}{G.~Francisco}, \bibinfo{author}{D.~Askari}, \bibinfo{author}{Y.~Yang},
\newblock \bibinfo{title}{Flow past a rotationally oscillating cylinder},
\newblock \bibinfo{journal}{Journal of Fluid Mechanics} \bibinfo{volume}{735} (\bibinfo{year}{2013}) \bibinfo{pages}{307–346}. \DOIprefix\doi{10.1017/jfm.2013.469}.
\bibitem[{Wang et~al.(2008)Wang, Fan, and Luo}]{WANG2008283}
\bibinfo{author}{Z.~Wang}, \bibinfo{author}{J.~Fan}, \bibinfo{author}{K.~Luo},
\newblock \bibinfo{title}{Combined multi-direct forcing and immersed boundary method for simulating flows with moving particles},
\newblock \bibinfo{journal}{International Journal of Multiphase Flow} \bibinfo{volume}{34} (\bibinfo{year}{2008}) \bibinfo{pages}{283--302}. \DOIprefix\doi{https://doi.org/10.1016/j.ijmultiphaseflow.2007.10.004}.
\bibitem[{Werbos(1990)}]{58337}
\bibinfo{author}{P.~Werbos},
\newblock \bibinfo{title}{Backpropagation through time: what it does and how to do it},
\newblock \bibinfo{journal}{Proceedings of the IEEE} \bibinfo{volume}{78} (\bibinfo{year}{1990}) \bibinfo{pages}{1550--1560}. \DOIprefix\doi{10.1109/5.58337}.
\bibitem[{Hasegawa et~al.(2020)Hasegawa, Fukami, Murata, and Fukagata}]{Hasegawa_2020}
\bibinfo{author}{K.~Hasegawa}, \bibinfo{author}{K.~Fukami}, \bibinfo{author}{T.~Murata}, \bibinfo{author}{K.~Fukagata},
\newblock \bibinfo{title}{Cnn-lstm based reduced order modeling of two-dimensional unsteady flows around a circular cylinder at different reynolds numbers},
\newblock \bibinfo{journal}{Fluid Dynamics Research} \bibinfo{volume}{52} (\bibinfo{year}{2020}) \bibinfo{pages}{065501}. \DOIprefix\doi{10.1088/1873-7005/abb91d}.
\bibitem[{Li et~al.(2024)Li, Liu, Peng, Yuan, and Wang}]{10.1063/5.0210493}
\bibinfo{author}{Z.~Li}, \bibinfo{author}{T.~Liu}, \bibinfo{author}{W.~Peng}, \bibinfo{author}{Z.~Yuan}, \bibinfo{author}{J.~Wang},
\newblock \bibinfo{title}{A transformer-based neural operator for large-eddy simulation of turbulence},
\newblock \bibinfo{journal}{Physics of Fluids} \bibinfo{volume}{36} (\bibinfo{year}{2024}) \bibinfo{pages}{065167}. \DOIprefix\doi{10.1063/5.0210493}.
\bibitem[{Shi et~al.(2026)Shi, Zhang, Sun, and Sui}]{10.1063/5.0312013}
\bibinfo{author}{C.~Shi}, \bibinfo{author}{G.~Zhang}, \bibinfo{author}{T.~Sun}, \bibinfo{author}{Y.~Sui},
\newblock \bibinfo{title}{Long-term prediction of cylinder water entry flow fields using a physics-informed spatiotemporal graph neural network},
\newblock \bibinfo{journal}{Physics of Fluids} \bibinfo{volume}{38} (\bibinfo{year}{2026}) \bibinfo{pages}{023327}. \DOIprefix\doi{10.1063/5.0312013}.
\bibitem[{Cui et~al.(2017)Cui, Yao, Wang, and Liu}]{CUI201724}
\bibinfo{author}{X.~Cui}, \bibinfo{author}{X.~Yao}, \bibinfo{author}{Z.~Wang}, \bibinfo{author}{M.~Liu},
\newblock \bibinfo{title}{A hybrid wavelet-based adaptive immersed boundary finite-difference lattice boltzmann method for two-dimensional fluid–structure interaction},
\newblock \bibinfo{journal}{Journal of Computational Physics} \bibinfo{volume}{333} (\bibinfo{year}{2017}) \bibinfo{pages}{24--48}. \URLprefix \url{https://www.sciencedirect.com/science/article/pii/S0021999116306726}. \DOIprefix\doi{https://doi.org/10.1016/j.jcp.2016.12.019}.
\bibitem[{Uhlmann(2005)}]{UHLMANN2005448}
\bibinfo{author}{M.~Uhlmann},
\newblock \bibinfo{title}{An immersed boundary method with direct forcing for the simulation of particulate flows},
\newblock \bibinfo{journal}{Journal of Computational Physics} \bibinfo{volume}{209} (\bibinfo{year}{2005}) \bibinfo{pages}{448--476}. \URLprefix \url{https://www.sciencedirect.com/science/article/pii/S0021999105001385}. \DOIprefix\doi{https://doi.org/10.1016/j.jcp.2005.03.017}.
\bibitem[{Wang et~al.(2015)Wang, Shu, Teo, and Wu}]{WANG2015440}
\bibinfo{author}{Y.~Wang}, \bibinfo{author}{C.~Shu}, \bibinfo{author}{C.~Teo}, \bibinfo{author}{J.~Wu},
\newblock \bibinfo{title}{An immersed boundary-lattice boltzmann flux solver and its applications to fluid–structure interaction problems},
\newblock \bibinfo{journal}{Journal of Fluids and Structures} \bibinfo{volume}{54} (\bibinfo{year}{2015}) \bibinfo{pages}{440--465}. \URLprefix \url{https://www.sciencedirect.com/science/article/pii/S0889974614002709}. \DOIprefix\doi{https://doi.org/10.1016/j.jfluidstructs.2014.12.003}.

\end{thebibliography}

\appendix
\setcounter{figure}{0}
\setcounter{table}{0}
\renewcommand{\thefigure}{\thesection\arabic{figure}}   
\renewcommand{\thetable}{\thesection\arabic{table}}
\renewcommand{\theHfigure}{\thesection\arabic{figure}}  
\renewcommand{\theHtable}{\thesection\arabic{table}}
\begin{appendices}
\section{Numerical verification} \label{A-}

A non-uniform orthogonal Cartesian mesh with $717\times382$ cells is employed over $\Omega_{f}$. Grid stretching is applied away from the body, while the mesh is locally refined around the cylinder within a rectangular region to resolve near-wall dynamics; the minimum spacing near the cylinder is $\Delta x_{\min}=\Delta y_{\min}=D/64.$ The immersed boundary is represented by $N_\mathrm{marker}=196$ uniformly distributed Lagrangian markers along the cylinder surface, and no-slip boundary conditions are imposed through direct forcing.

\begin{table}[width=.8\linewidth,cols=2,pos=h]
\caption{\label{table001}Dimensionless coefficients obtained from the simulation of the flow around a stationary cylinder at $Re=100$, using $D=1$ and $dt=0.005$.}

\begin{tabular*}{\tblwidth}{@{} LLLLL @{}}
\toprule
 & $\bar{C}_D$ & $C_L'$ & $St$ \\
\midrule
Our & 1.383 & $\pm 0.345$ & 0.167 \\
Cui et al.\cite{CUI201724} & 1.360 & $\pm 0.340$ & 0.167 \\
Uhlmann\cite{UHLMANN2005448} & 1.453 & $\pm 0.339$ & 0.169 \\
Wang et al.\cite{WANG2015440} & 1.334 & $\pm 0.370$ & 0.163 \\
\bottomrule
\end{tabular*}

\end{table}

\section{Training details of neural model} \label{A}
\subsection{Downsampling of high-resolution ground-truth data} \label{A.1}

\begin{figure}[htbp]
\centering
\includegraphics[scale=.9]{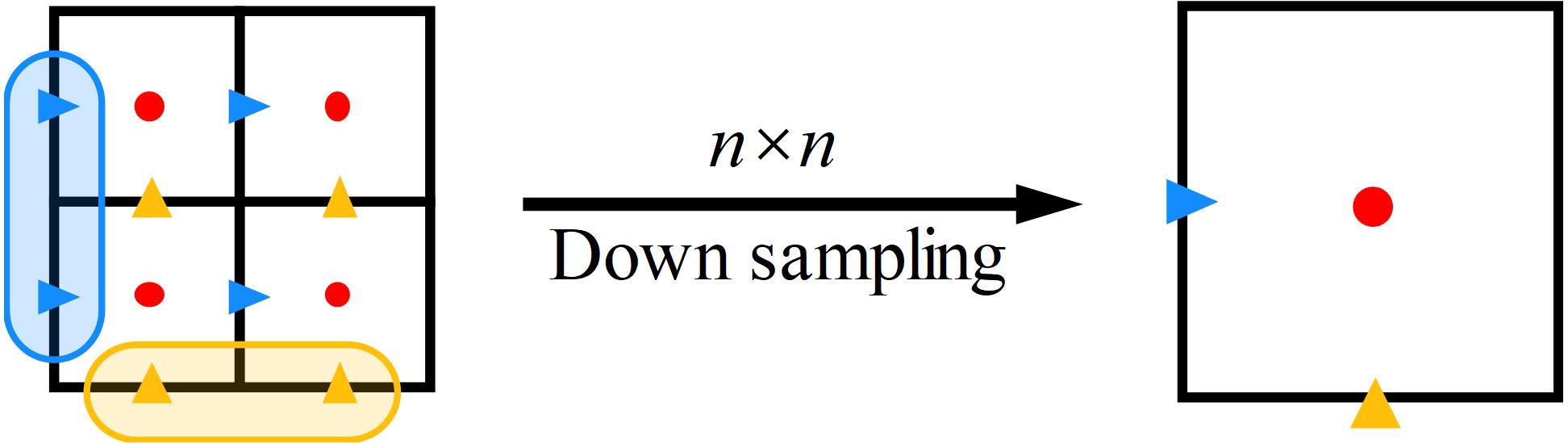}
\caption{Illustration of transferring staggered-grid variables from a fine mesh to an $n \times n$ coarser mesh via downsampling (an $8\times$ reduction in resolution is used in this study). For each coarse control volume, the face-centered velocity components on the fine grid are area-averaged to obtain the corresponding coarse-grid velocities, while the cell-centered pressures within the fine-grid subcells are averaged to yield the coarse-grid pressure. Fine-grid velocity samples located in the interior of the coarse cell are not retained.}
\label{downsample}
\end{figure}

The training dataset is constructed from the high-fidelity simulation data through two successive steps. First, in the spatial dimension, all field variables are transferred from the fine grid ($717 \times 382$ cells over $\Omega_f$) to the coarse surrogate grid ($80 \times 40$ cells over $\Omega^r_f$, corresponding to an $8\times$ reduction in resolution) via the downsampling procedure illustrated in Fig.~\ref{downsample}: face-centered velocity components are area-averaged over each coarse control volume, and cell-centered pressures are averaged over the corresponding fine-grid subcells. Second, in the temporal dimension, the spatially downsampled flow fields and structural forces are sampled at every 100 high-fidelity time steps, yielding a learning time step of $\Delta t = 100 \times dt = 0.5$. This sampling interval directly determines $\Delta t$, which defines the prediction horizon of the surrogate model (see Eq.~\eqref{eq:intermediate velocity}). 

\subsection{Hyperparameters of the trainable ConvResNet block} \label{A.2}
Each ConvResNet block is composed of five convolutional layers with channel configuration $[32, 64, 64, 32, 32]$ and a trainable $3 \times 3$ kernel. A rectified linear unit (ReLU) is applied after each layer except for the last layer, where no activation is used to preserve the linearity of the output fields.

The initial learning rate is set to $10^{-3}$, and the Adam optimizer is adopted with a weight decay of $10^{-5}$. A ReduceLROnPlateau learning-rate scheduler is used, configured with mode = min, factor = 0.5, patience = 2, cooldown = 0, and min\_lr = $10^{-6}$. The model is trained for 3000 epochs with a batch size of 128.

In our implementation, the learning-rate scheduler is driven by the validation loss rather than the training loss. This choice is motivated by the intended deployment mode of the proposed model: it is ultimately used for autoregressive rollouts, where prediction errors accumulate over time and the effective performance is governed by generalization under distribution shift induced by the model itself. Therefore, using the validation loss as the control signal provides a more reliable indicator of rollout-ready model quality than the one-step training objective alone, and yields a more sensible criterion for adapting the learning rate during optimization.

\begin{figure}[htbp]
\centering
\includegraphics[scale=.65]{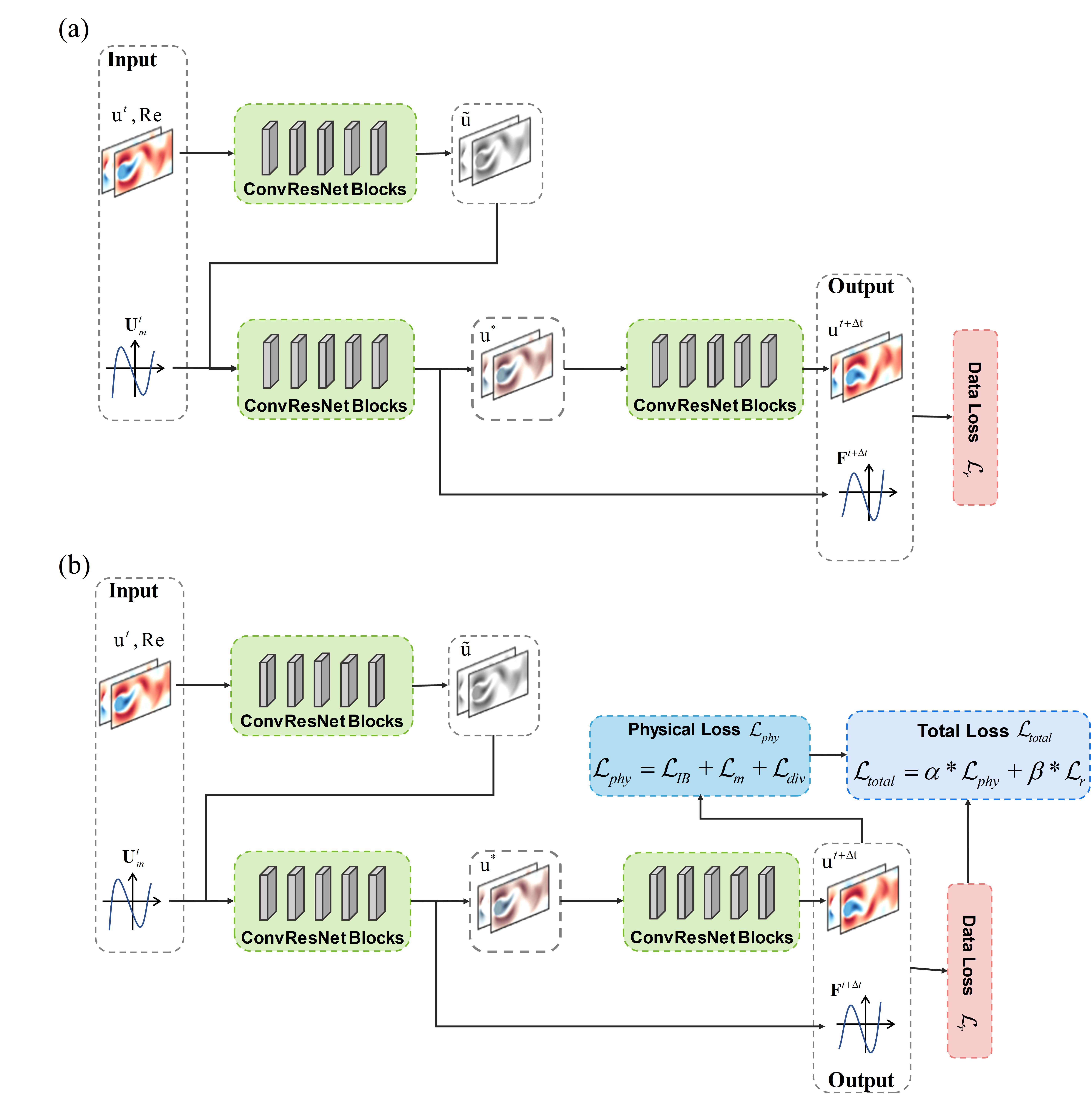}
\caption{Architectures of the baseline models used for comparison. (a) Purely data-driven model trained with the data loss $\mathcal{L}_r$. (b) Physics loss constrained model, which augments (a) with the physics-consistency term $\mathcal{L}_{\mathrm{phy}} = \mathcal{L}_{\mathrm{IB}} + \mathcal{L}_m + \mathcal{L}_{\mathrm{div}}$ and optimizes the total loss $\mathcal{L}_{\mathrm{total}} = \alpha \mathcal{L}_{\mathrm{phy}} + \beta \mathcal{L}_r$.}
\label{baseline model}
\end{figure}

\section{Architecture of baseline models(Fig.~\ref{baseline model})} \label{B}

\end{appendices}

\end{document}